\documentclass[10pt,twocolumn,letterpaper]{article}

\usepackage{wacv}      %

\usepackage{newtxtext,newtxmath}

\usepackage{amsmath}
\usepackage{booktabs}
\usepackage{multirow}
\usepackage{graphicx}
\usepackage{bbm}
\usepackage{float}
\usepackage{xcolor}
\usepackage{enumitem}

\usepackage{arydshln}
\usepackage{wasysym}
\newcommand{\Inf}{\mathrm{Inf}}
\newcommand{\method}{\mbox{HullProbe}}
\newcommand{\releaseurl}{\url{https://github.com/fossil-ai/HullProbe}}
\newcommand{\cmark}{{\footnotesize\CIRCLE}}       %
\newcommand{\pmark}{{\footnotesize\LEFTcircle}}   %
\newcommand{\xmark}{{\footnotesize\Circle}}       %
\newcommand{\namark}{\textcolor{black!45}{--}}    %

\definecolor{wacvblue}{rgb}{0.21,0.49,0.74}
\usepackage[breaklinks,colorlinks,allcolors=wacvblue]{hyperref}
\hypersetup{pdftitle={HullProbe: Measuring Interventional Part Reliance in Maritime Vessel Classifiers},
  pdfauthor={Faisal Mohammad, Mohamed Younis}}

\def\wacvPaperID{1505}
\def\confName{WACV}
\def\confYear{2027}

\title{\method: Measuring Interventional Part Reliance in Maritime Vessel Classifiers}

\author{Faisal Mohammad$^{1,2}$ \qquad Mohamed Younis$^{1}$\\
$^{1}$University of Maryland, Baltimore County\\
$^{2}$Johns Hopkins University Applied Physics Laboratory\\
{\tt\small faisalm2@umbc.edu, faisal.mohammad@jhuapl.edu, younis@umbc.edu}
}

\newcommand{\ieeenotice}{\AddToShipoutPicture*{\AtPageLowerLeft{\put(\LenToUnit{0.5\paperwidth},\LenToUnit{0.42in}){\makebox[0pt]{\parbox{0.84\paperwidth}{\centering\scriptsize
\copyright\ 2027 IEEE. Personal use of this material is permitted. Permission from IEEE must be obtained for all other uses, in any current or future media, including reprinting/republishing this material for advertising or promotional purposes, creating new collective works, for resale or redistribution to servers or lists, or reuse of any copyrighted component of this work in other works.}}}}}}

\begin{document}
\ieeenotice
\raggedbottom
\maketitle
\begin{abstract}
Recognizing a vessel's type from an image supports maritime surveillance and sanctions monitoring,
but a classifier can reach high accuracy by leaning on the generic hull instead of the more
distinctive superstructure, and is then fooled when a vessel is re-hulled or occluded. De-biasing
methods claim to remove such reliance, yet the field certifies them with worst-group accuracy, a
proxy that needs group labels and only infers reliance from accuracy. We measure the reliance directly. \method\ swaps a registered anatomical part of a vessel
for another instance's and reads how much that part drives the decision, using a same-class swap to
control for compositing artifacts. Because a surface vessel is a hull and a superstructure arranged
along one axis, its parts form a $2{\times}3$ grid whose $2^6{=}64$ swaps we evaluate exhaustively,
and a single Walsh transform returns each part's interventional influence and its interaction
spectrum. On planted shortcuts, where the true reliance is known, the measure tracks worst-group
accuracy ($r{=}{-}0.94$ over $45$ configurations), and replacing its learned segmenter with human
annotations changes the hull-share by less than the audit's resolution. On the $26$ vessel types of MARVEL,
the classifier places about $46\%$ of its corrected sensitivity on the generic hull yet follows the
superstructure when the two conflict, and the group-label-free de-biasers available there do not
detectably reduce the hull's share. The findings hold across three architectures and a second
maritime dataset, and the method extends to fine-grained vehicle recognition on Stanford Cars.
\end{abstract}

\section{Introduction}
\begin{figure}[t]
\centering
\includegraphics[width=0.85\linewidth]{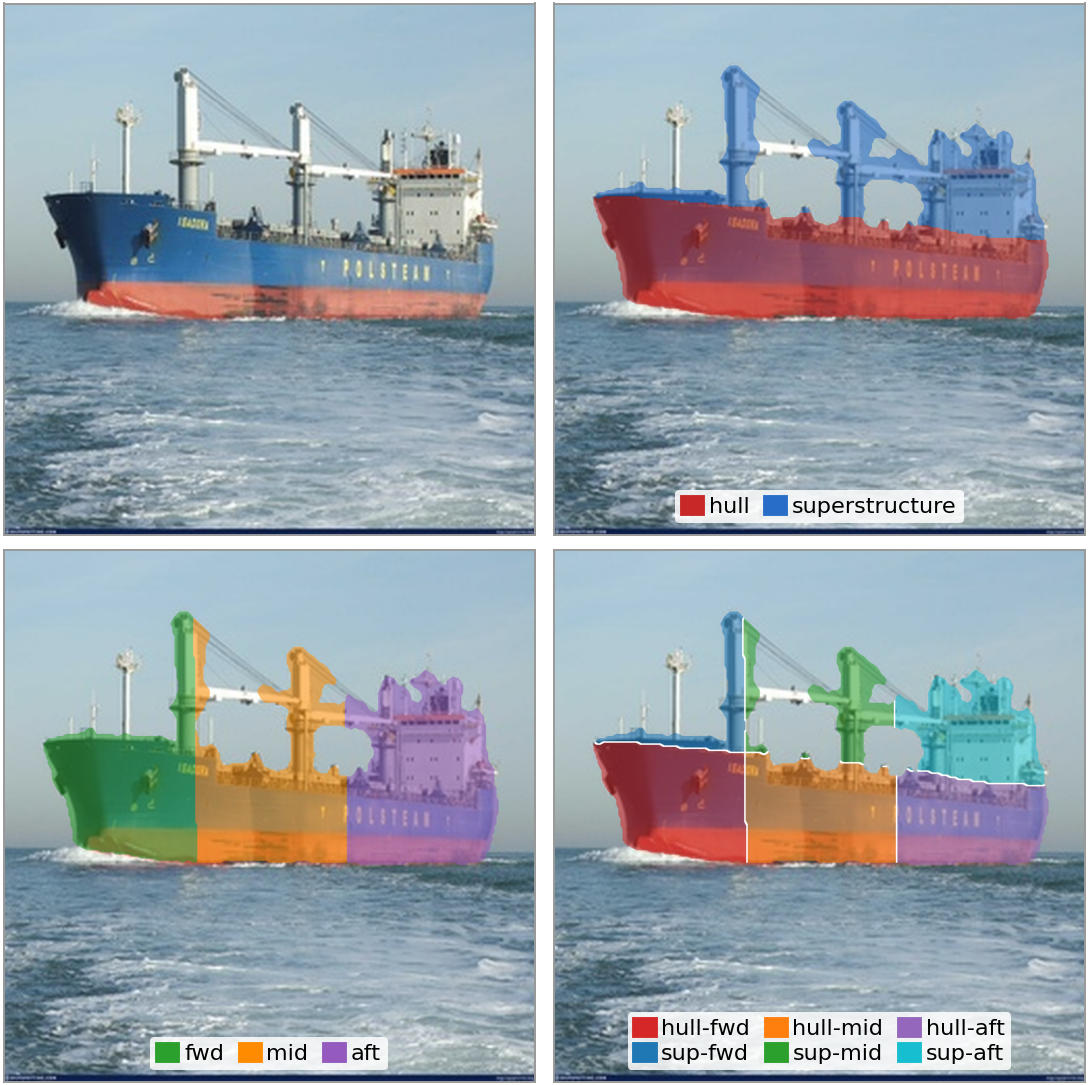}
\caption{\textbf{Two orthogonal decompositions give a $2{\times}3$ part grid.} From a vessel
(top-left), a learned segmenter splits it vertically into \{hull, superstructure\} (top-right)
and its PCA pose axis splits it longitudinally into canonical \{forward, mid, aft\} thirds
(bottom-left); their product is a grid of six anatomically defined cells (bottom-right, each named
in its legend), each individually registrable and swappable for a donor's.}
\label{fig:decomp}
\end{figure}
A ship's type is defined mostly by its superstructure: the deck cranes of a bulk carrier, the
stacked cells of a container ship, and the deck piping of a tanker distinguish types that otherwise
share a broadly similar hull. A classifier can still reach high accuracy by leaning on hull cues that
merely correlate with type in the training data~\cite{shortcut,texturebias,backgrounds}, rather than
on the superstructure that determines it. Such a model is brittle where it matters, in maritime
surveillance and sanctions monitoring, failing when a vessel is re-hulled or when its superstructure
is occluded and only the hull remains. A substantial literature attacks this reliance on a spurious
cue under group robustness and de-biasing: GroupDRO~\cite{groupdro}, JTT~\cite{jtt},
LfF~\cite{lff}, DFR~\cite{dfr}, and simple group balancing~\cite{subg} all aim to reduce a
classifier's dependence on a spurious feature.

\begin{table*}[t]
\centering
\caption{\textbf{Where HullProbe sits among approaches to measuring feature reliance}
(Sec.~\ref{sec:related}). Columns are the five properties discussed in Sec.~\ref{sec:related};
\cmark~has it, \pmark~partial, \xmark~lacks it, \namark~not applicable. \emph{Exact} = exhaustive
over the $64$-point intervention set, not sampled over subsets (the donor law is still estimated).}
\label{tab:positioning}
\renewcommand{\arraystretch}{1.2}
\begin{tabular}{@{}l ccccc@{}}
\toprule
Approach & Direct & Interventional & Plausible & Group-label-free & Exact \\
\midrule
Worst-group accuracy~\cite{groupdro}                  & \xmark~{\scriptsize(proxy)} & \xmark & \namark & \xmark & \namark \\
\hdashline
Gradient saliency (Grad-CAM~\cite{gradcam})           & \xmark & \xmark & \namark & \cmark & \namark \\
\hdashline
Occlusion / removal (RISE~\cite{rise}, ROAR~\cite{roar}, SAGE~\cite{sage}) & \pmark & \cmark & \xmark & \cmark & \xmark \\
\hdashline
Counterfactual / generative edits~\cite{goyal_cf,chang_cf,sauer_cgn} & \pmark & \cmark & \cmark & \cmark & \xmark \\
\hdashline
Functional decomposition (Shapley~\cite{shap}, Sobol~\cite{sobol}) & \cmark & \pmark & \xmark & \cmark & \xmark \\
\midrule
\textbf{HullProbe (ours)}                             & \cmark & \cmark & \pmark & \cmark & \cmark~{\scriptsize(64)} \\
\bottomrule
\end{tabular}
\end{table*}

The standard way to certify that these methods work is \emph{worst-group
accuracy} (WGA)~\cite{groupdro}, the accuracy on the minority group in which the shortcut does not hold,
and the standard yardstick across group-robustness benchmarks~\cite{wilds,subpop}. If it
rises, the shortcut is declared broken. This proxy is indirect, inferring reduced reliance from
improved accuracy on the hard cases even though the two need not coincide, since a model can raise
minority accuracy by better calibration or memorization while its reliance is unchanged. Whether
de-biasing reduces the reliance itself is, to our knowledge, not directly tested; measuring it
cleanly is hard, since correlational attribution and off-manifold occlusion give no clean
interventional reading and the standard benchmarks are synthetic (Sec.~\ref{sec:related}).

We contribute \textbf{HullProbe}, an evaluation method that estimates \emph{anatomical interventional
sensitivity} by performing an anatomically constrained part replacement on a real vessel and reading how much
swapping one component moves a frozen classifier's decision. This sensitivity is easy to confuse with two
other things, and we keep it apart from both. One is \emph{sufficiency}~\cite{watson_ns}, whether a part on its own
sets the predicted class (Sec.~\ref{sec:hijack}); the other is \emph{spuriousness}~\cite{shortcut,groupdro}, whether relying
on the part is a genuine shortcut, which we can only establish for the controlled shortcuts
(Sec.~\ref{sec:phase2}). On the natural task spuriousness stays open, and HullProbe reports where
sensitivity lives, not whether it ought to be removed. The maritime domain makes this
possible, because a surface vessel admits two orthogonal, exhaustive part decompositions that
register into one canonical frame. A learned segmenter splits the vessel vertically into \{hull,
superstructure\}; its long axis, found by principal component analysis of the vessel mask (its
pose), splits it longitudinally into \{forward, mid, aft\}. Their
product is a $2{\times}3$ grid of six anatomical cells (Fig.~\ref{fig:decomp}). Replacing
any cell with a donor vessel's registered cell produces an anatomically plausible ship; writing the replaced
cells as a bit vector $z\in\{0,1\}^6$, the classifier's behavior over all $64$ configurations
is a function on a Boolean cube that we evaluate exhaustively. This recovers the
decomposition of class evidence into single-cell effects and interactions as an exact functional
ANOVA over the constructed cube (the donor expectation is estimated by Monte-Carlo, Sec.~\ref{sec:method}).

\noindent\textbf{Contributions.}
\begin{itemize}[leftmargin=1.2em,itemsep=1pt,topsep=1pt]
\item \textbf{A direct, group-label-free measure of part reliance.} An exact functional ANOVA over
all $64$ registered part swaps, with a same-class artifact control and a semantic-$L_0$ radius
(Sec.~\ref{sec:method}).
\item \textbf{Validation.} On planted shortcuts with known ground truth the measure tracks
worst-group accuracy ($r{=}{-}0.94$ over $45$ configurations, up to $-0.97$ across shortcuts), and
swapping its learned segmenter for human annotations moves the hull-share by only $-0.01$,
$95\%$ CI $[-0.04,0.02]$ (Sec.~\ref{sec:validation}).
\item \textbf{A finding about de-biasing.} In controlled shortcuts, group-supervised methods (e.g.\
DFR) sharply reduce measured reliance; on natural MARVEL data, none of the tested group-label-free
methods reduces relative hull-share beyond the audit's ${\sim}0.04$ resolution
(Secs.~\ref{sec:phase2},~\ref{sec:debias}).
\end{itemize}
\noindent Sec.~\ref{sec:validation} validates the instrument before Sec.~\ref{sec:results} applies it.

\begin{figure*}[t]
\centering
\includegraphics[width=\linewidth]{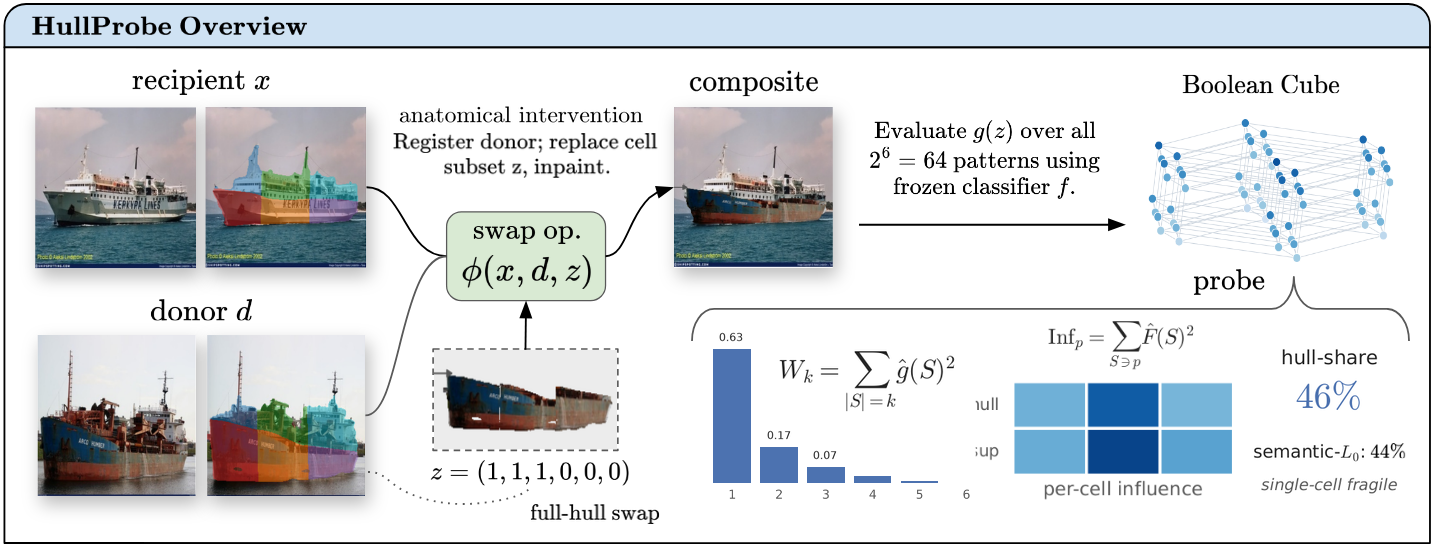}
\caption{\textbf{HullProbe overview.} A recipient $x$ and donor $d$ are decomposed into a
$2{\times}3$ grid of six anatomical cells (hull/superstructure $\times$ forward/mid/aft). The
swap operator $\phi(x,d,z)$ replaces any cell subset $z\in\{0,1\}^6$ of the recipient with the
donor's registered cells (here a full-hull swap, $z{=}(1,1,1,0,0,0)$), producing a plausible
composite. Evaluating the frozen classifier $f$ over all $2^6{=}64$ patterns turns its
behavior into a function $g(z)$ on the Boolean cube; one exact Walsh--Hadamard transform then
yields the functional-ANOVA spectrum $W_k$, the per-cell influence $\mathrm{Inf}_p$ and its
hull-share ($46\%$), and the semantic-$L_0$ radius.}
\label{fig:arch}
\end{figure*}

\section{Related Work}
\label{sec:related}
\noindent\textbf{Shortcut learning and spurious correlations.} Networks preferentially exploit
easy, spurious cues such as texture over shape~\cite{texturebias} or background and geography
over object~\cite{backgrounds,terra}, and more broadly any feature correlated with the label in
training~\cite{shortcut,salientimagenet}; shedding such cues is a stated goal of causal
representation learning~\cite{causal_rep}. Our
target is one specific, physically meaningful instance, the reliance on the vessel hull
rather than the class-defining superstructure.

\smallskip\noindent\textbf{Group robustness / de-biasing.} Methods to reduce spurious reliance
divide into two families. Group-supervised methods use the group labels: GroupDRO~\cite{groupdro}
optimizes worst-group loss, subsampling/reweighting~\cite{subg} is a strong cheap baseline, and
DFR~\cite{dfr} retrains only the last layer on group-balanced data. Group-free methods infer the
hard cases: JTT~\cite{jtt} upweights the errors of an ERM model, LfF~\cite{lff} trains against a
deliberately biased model, and AFR~\cite{afr} reweights features automatically; a parallel line
enforces invariance across inferred environments or subclasses~\cite{irm,eiil,cnc,george}, and
others constrain where the model looks, through masking~\cite{masktune} or attribution
priors~\cite{ross_rrr}. All are evaluated by worst-group accuracy on a small set of
benchmarks~\cite{wilds,subpop}, several of them (e.g.\ Waterbirds~\cite{groupdro}) built by
segment-and-composite, the construction our planted shortcuts (Sec.~\ref{sec:phase2}) share but use
diagnostically. We do
not propose a new such method; we give an interventional readout of what they do to part reliance,
which the accuracy proxy only infers.

\smallskip\noindent\textbf{Counterfactual and interventional attribution.} Interventional attribution
changes a feature and measures the output change, in the interventional sense of~\cite{pearl,janzing} and the
counterfactual-invariance view of spurious correlations~\cite{veitch}, unlike
correlational saliency such as attribution maps (Grad-CAM~\cite{gradcam},
LIME~\cite{lime}, SmoothGrad~\cite{smoothgrad}, DeepLIFT~\cite{deeplift}) and concept or unit
probes~\cite{tcav,netdissect,protopnet}, which can be unstable or fail sanity checks~\cite{fragile,sanity}.
Its recurring difficulty is getting a clean intervention, since masking or occlusion is off-manifold
and confounded by covariate shift~\cite{meaningful_pert,rise}, while generative counterfactual
edits~\cite{goyal_cf,chang_cf,sauer_cgn} are realistic but hard to localize to a single part. HullProbe
avoids both, because its swaps stay plausible and target a named part while remaining
artifact-controlled. Because the space is only $64$ points, HullProbe evaluates it completely and
recovers all interactions.

\smallskip\noindent\textbf{Functional decompositions.} Shapley attribution~\cite{shap} and functional
ANOVA / Sobol indices~\cite{efron_stein,sobol,owen_shapley} decompose a prediction into feature
contributions and interactions but are estimated by sampling over exponentially many
subsets. On our six-cell cube the decomposition is computed exactly through the
Fourier--Walsh expansion, whose influences and low-degree concentration are standard
objects in the analysis of Boolean functions~\cite{lmn,odonnell}.

\smallskip\noindent\textbf{Region mixing and adversarial edits.} CutMix~\cite{cutmix},
MixUp~\cite{mixup}, and broader augmentation~\cite{cutout,randaugment,augmix,lisa} blend inputs as
regularizers, which we audit rather than use. Adversarial perturbations~\cite{fgsm,pgd,advpatch,jsma}
are signal-level; our edit is a semantic, anatomically localized part swap used diagnostically.

\noindent Among these approaches (Table~\ref{tab:positioning}), HullProbe is alone in being at once
direct, interventional, group-label-free, and exact over the constructed cube; the partial
plausibility mark reflects that the composites, though anatomically plausible, are not photorealistic
and not shown to lie on the natural-image manifold.

\section{Methodology}
\label{sec:method}
HullProbe measures how much a frozen classifier's decision depends on each anatomical part,
without modifying it (Fig.~\ref{fig:arch}).

\subsection{Two decompositions and the part grid}
The vessel silhouette is the boat/ship mask of an ADE20K-pretrained
SegFormer-B2~\cite{segformer,ade20k}, used zero-shot. To split it, we fine-tune the same model on
$1{,}014$ vessels whose hull/superstructure boundary an annotator placed (App.~\ref{sec:operator-appendix};
a promptable or instance segmenter~\cite{sam,maskrcnn} is an alternative), reaching mean
hull/superstructure IoU $0.756$ on held-out vessels, and apply it to all $34{,}768$ vessels of
MARVEL~\cite{marvel} subset~B. Inside the silhouette, the hull is everything the segmenter does not
label superstructure, so the two are touching and jointly complete.
Principal component analysis of the vessel mask gives its long axis, which we call the
pose; splitting along it partitions the vessel longitudinally into forward / mid / aft thirds. The product is the cell set
$P=\{\text{hull},\text{sup}\}\times\{\text{fwd},\text{mid},\text{aft}\}$, $|P|{=}6$,
which exactly partitions the vessel (Fig.~\ref{fig:decomp}).

\subsection{Registered anatomical cell intervention}
Recipient $x$ (true label $y$) and donor $d$ are registered into a common canonical frame by
a similarity transform (derotate by pose, normalize length, and align the deck). For a subset
$z\in\{0,1\}^6$ ($z_p{=}1$ if cell $p$ is taken from the donor, $0$ if kept), we fit each
swapped donor cell to the recipient's cell (canonical affine on longitudinal extent, area,
centroid) and stencil it to the recipient's exact cell support. The silhouette is then the
recipient's, with no overhang or holes; retained cells stay pixel-exact, and any residual is
inpainted (full mechanics in App.~\ref{sec:operator-appendix}). Denote the result
$\phi(x,d,z)$: $z{=}\mathbf 0$ is the original and $z{=}(1,1,1,0,0,0)$ a full hull swap.
Cross-class donors are chosen by aspect-ratio compatibility, isolating a class change under
matched shape; the output remains a plausible vessel.

\begin{figure}[t]
\centering
\includegraphics[width=\linewidth]{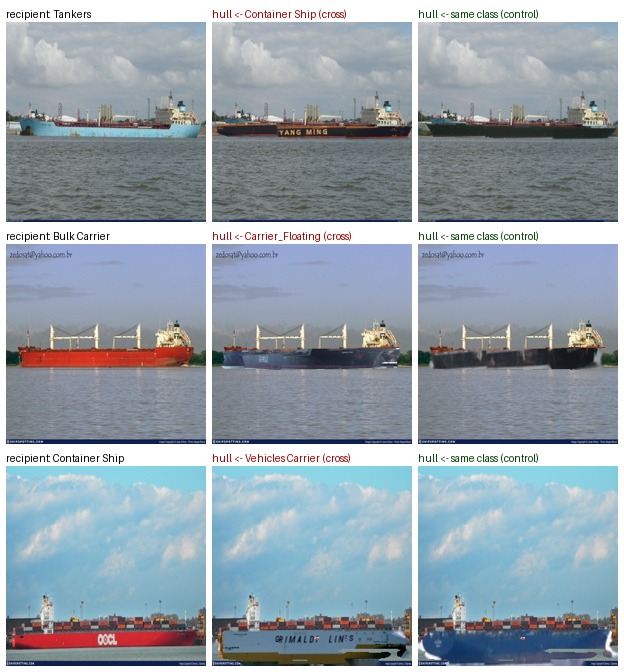}
\caption{\textbf{Operator and control.} For a recipient (left), a cross-class
hull swap (middle) changes the hull's class while keeping the superstructure; the
same-class swap (right) carries the identical seams/inpainting but preserves the
label, and is the artifact control subtracted in Eq.~\eqref{eq:corr}. Both are
anatomically plausible vessels, though not photorealistic.}
\label{fig:qual}
\end{figure}

\subsection{Class evidence on the cube: effects and interactions}
We first summarize how the frozen classifier behaves across swap patterns. For a fixed recipient,
and to separate the effect of which cells are swapped from which donor supplies them,
we average over a donor law $\pi$,
\begin{equation}
g(z) \;=\; \mathbb{E}_{d\sim\pi}\big[\,\mathbf 1\{\arg\max_k f_k(\phi(x,d,z))=y\}\,\big],
\label{eq:g}
\end{equation}
the rate at which $f$ still predicts $y$ under swap pattern $z$; by construction $g$ takes values in
$[0,1]$. Because $z$ has only $64$ values, $g:\{0,1\}^6\!\to\![0,1]$ is evaluated everywhere.

Two readouts follow directly from $g$, without any transform. The \emph{main effect} of
cell $p$ is the average drop in class evidence from toggling that cell, marginalizing
the other five:
\begin{equation}
M_p \;=\; \mathbb{E}_{z_{-p}}\big[\,g(z_{-p},z_p{=}0)-g(z_{-p},z_p{=}1)\,\big],
\label{eq:main}
\end{equation}
where $z_{-p}$ denotes the other five bits. A large $M_p$ means swapping cell $p$
reliably destroys the correct decision. The pairwise interaction of cells
$p,q$ is the extent to which their joint effect departs from additivity,
\begin{equation}
I_{pq}=\mathbb{E}_{z_{-pq}}\big[g_{00}-g_{10}-g_{01}+g_{11}\big],
\label{eq:inter}
\end{equation}
with $g_{ab}$ shorthand for $g$ at $z_p{=}a,z_q{=}b$ (others fixed).

These two readouts are the degree-$1$ and degree-$2$ components of $g$: up to a constant they
are its first- and second-order Fourier--Walsh coefficients ($\hat g(\{p\}){=}\tfrac12 M_p$,
$\hat g(\{p,q\}){=}\tfrac14 I_{pq}$; App.~\ref{sec:fw-appendix}). Since $g$ is defined on all
$64$ configurations, summing their squared magnitudes by order is an exact functional ANOVA:
$\operatorname{Var}_z[g]$ splits with no residual into a degree-$1$ part from the $M_p$, a
degree-$2$ part from the $I_{pq}$, and higher orders, via the orthonormal parity
(Fourier--Walsh) basis~\cite{lmn,odonnell}. We write $W_k$ for the fraction at order $k$, so
$W_1$ and $W_2$ are exactly the energy of the main effects and pairwise interactions above;
concentration at low $k$ means the response function $g$ is largely additive over cells, so a per-cell
attribution captures most of its structure. We report this $W_k$ spectrum
(Fig.~\ref{fig:spectrum}a) and defer the mechanics to App.~\ref{sec:fw-appendix}.

\subsection{Influence and hull-share}
For the decision itself we use the hard counterpart of the soft main effect $M_p$. For a donor $d$,
let $F_d(z)\in\{-1,+1\}$ be $+1$ iff $f$ predicts $y$ on
$\phi(x,d,z)$. The influence of cell $p$ is the probability that toggling it
flips the decision, averaged over the donor law $\pi$,
\begin{equation}
\Inf_p \;=\; \mathbb{E}_{d\sim\pi}\,\Pr_{z}\big[F_d(z)\ne F_d(z\oplus e_p)\big],
\label{eq:influence}
\end{equation}
where $z\oplus e_p$ flips bit $p$. Equivalently $\Inf_p=\mathbb{E}_{d}\sum_{S\ni p}\hat F_d(S)^2$, the
fraction of $F_d$'s energy routed through coordinate $p$ (App.~\ref{sec:fw-appendix}). Our
summary statistic is the share of total influence carried by the three hull cells,
\begin{equation}
\text{hull-share}\;=\;\frac{\sum_{p\in\text{hull}}\Inf_p}{\sum_{p\in P}\Inf_p},
\label{eq:hullshare}
\end{equation}
what fraction of the model's interventional sensitivity lives in the generic hull. The \emph{raw}
hull-share, formed from the influences $\Inf_p$, lies in $[0,1]$; every reported value is the
\emph{corrected} hull-share, formed from the artifact-corrected $\Inf_p^{\text{corr}}$
(Eq.~\eqref{eq:corr}). As a difference of two influences $\Inf_p^{\text{corr}}$ can be marginally
negative for a class-agnostic cell (e.g.\ car wheels, App.~\ref{app:cars}), so the corrected share is
not strictly bounded, though in practice it stays well inside $[0,1]$.

\subsection{Semantic-$L_0$ radius}
The worst-case counterpart to average influence is the fewest cells whose swap flips the
class with some realistic donor,
\begin{equation}
r(x)=\min\{\,|S| : \exists\,d,\ \arg\max_k f_k(\phi(x,d,\mathbf 1_S))\ne y\,\},
\label{eq:l0}
\end{equation}
an anatomically constrained, semantic analogue of the $L_0$ adversarial radius~\cite{jsma,onepixel}
and of worst-case robustness to interventions~\cite{interv_robust} ($\mathbf 1_S$ is the
indicator of $S$). It is computed over all $63$ non-empty subsets against a fixed donor
bank, so it is exact only relative to that bank.

\subsection{Artifact control}
Because each swap stencils only the donor's cell texture onto the recipient's exact cell
support, the recipient's background, water, and scale are retained (scale is normalized when the vessel is fit
into the canonical frame), so those cues are not donor-injected. What remains is the donor cell's
class-identifying appearance, which is the intended signal, plus compositing artifacts that confound the
estimator of Eqs.~\eqref{eq:influence}--\eqref{eq:l0}, since seams and inpainting
can flip predictions independently of the donor cell's class. We adjust for this with a
control condition (Fig.~\ref{fig:qual}), the identical swap using a same-class donor,
which preserves the label but carries the same artifacts.
Writing $\Inf_p^{\times}$ (cross-class) and $\Inf_p^{=}$ (same-class), the
artifact-corrected influence is
\begin{equation}
\Inf_p^{\text{corr}}=\Inf_p^{\times}-\Inf_p^{=}.
\label{eq:corr}
\end{equation}
If the artifact effect is class-independent (same compositor and registration, so same- and
cross-class artifact severities are comparable), Eq.~\eqref{eq:corr} is a paired subtraction that
adjusts for it. It is an artifact-adjusted estimate under that assumption, not a guaranteed
cancellation: $\Inf_p^{\text{corr}}$ estimates the classifier's interventional sensitivity to a
part's class-identifying appearance, not a model-free causal contribution of the part. The
convergent validity of Sec.~\ref{sec:phase2} uses the raw flip rate and does not depend on the
assumption, which Sec.~\ref{sec:sensitivity} probes directly. The same-class floor is large
(Sec.~\ref{sec:probe}), so the control is necessary.

\subsection{Estimation and cost}
We estimate $g,F$ by Monte-Carlo over $K_{\times}{=}4$ cross-class donors and $K_{=}{=}2$ same-class
controls per recipient (App.~\ref{app:protocol}); the $64{\times}(K_{\times}{+}K_{=})$ forward passes
per recipient are shared across audited classifiers, as composites are model-independent. The
transform is a $64$-point Walsh--Hadamard product. We aggregate per-recipient quantities with paired
bootstrap and, for the de-biasing audit, across-seed mean$\pm$std.

\section{Validating HullProbe}
\label{sec:validation}
Before applying HullProbe we test it as an instrument: does it recover reliance where the truth is
known by construction (Sec.~\ref{sec:phase2}), and how far do its own choices move the number
(Sec.~\ref{sec:sensitivity})?

\subsection{Setup}
\label{sec:setup}
\noindent\textbf{Data.} MARVEL~\cite{marvel} subset~B, $26$ vessel types (other maritime sets
are smaller-vocabulary or detection-oriented~\cite{mcships,smd,seaships}). Recipients and donors
are drawn from the held-out test split, disjoint from any training donor. Each analysis draws a
class-balanced recipient sample, scored over $K$ cross-class donors and same-class controls; we
state each population.

\smallskip\noindent\textbf{Classifiers.} The controlled-shortcut studies fine-tune
ResNet-50~\cite{resnet} models (three seeds) under five de-biasers, from ERM to the group-supervised
GroupDRO~\cite{groupdro} and DFR~\cite{dfr}. The natural-task analyses use a
ConvNeXt-Tiny~\cite{convnext} at $77.6\%$ test accuracy, and the de-biasing audit compares seven
ConvNeXt-Tiny models under a matched protocol (three seeds each): ERM, CutMix, hull-randomization,
background-swap, JTT~\cite{jtt}, LfF~\cite{lff}, and AFR~\cite{afr}
(App.~\ref{sec:methods-appendix}).

\begin{figure*}[t]\centering
\includegraphics[width=0.86\linewidth]{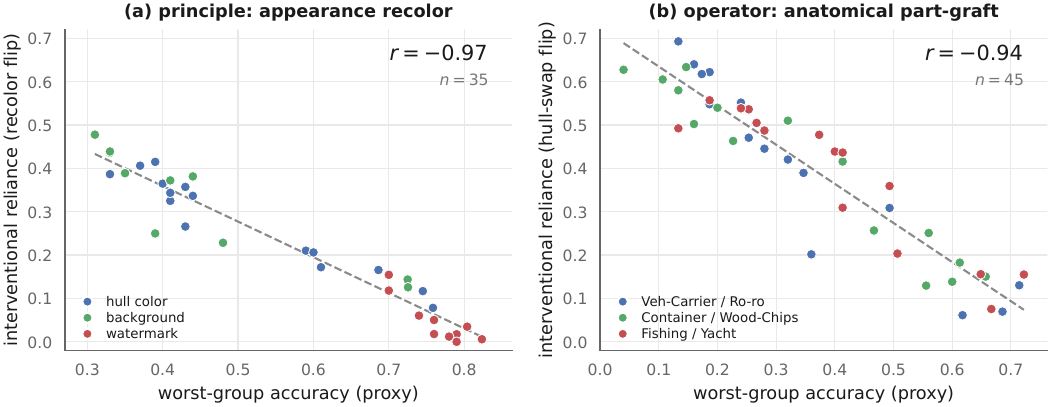}
\caption{\textbf{Convergent validity on planted shortcuts} (Sec.~\ref{sec:phase2}).
Interventional reliance vs.\ worst-group accuracy, colored by shortcut. \emph{(a)} The measurement
\emph{principle:} three appearance recolors (hull color, background, watermark) track WGA (pooled
$r{=}{-}0.97$, $35$ configs). \emph{(b)} The part-swap \emph{operator itself:} three anatomical
hull-grafts are planted and measured with the real $6$-cell swap, tracking WGA across $45$ configs
($3$ shortcuts $\times$ $3$ seeds $\times$ $5$ de-biasers; pooled $r{=}{-}0.941$). Each point is one
de-biaser$\times$seed; the dashed line is the least-squares fit.}
\label{fig:validation}
\end{figure*}

\subsection{Convergent validity on planted shortcuts}
\label{sec:phase2}
On natural data the true reliance is unknown and worst-group accuracy cannot be computed, so we
validate where both are fixed by construction. We plant a shortcut in a known location, the hull,
make ERM learn it through a $90\%$ train-time correlation with a binary label whose true cue is the
superstructure (balanced at test), and remove it to graded degrees with five de-biasers, from none
(ERM) to group-supervised (DFR). We therefore know where the shortcut is, that ERM uses it, and
which methods target it, and we validate at two levels.

For the measurement \emph{principle}, we tint the hull region red or blue. The \emph{recolor-flip
rate} (fraction of correct images whose prediction flips when the hull is recolored) tracks
worst-group accuracy at pooled $r{=}{-}0.97$ across five de-biasers and three recolor shortcuts
($35$ configs; Fig.~\ref{fig:validation}a, Table~\ref{tab:replication}). A watermark too small to
be learned serves as a known null, and the measure correctly reads near zero for it.

For the \emph{operator itself}, we plant the shortcut with the actual $6$-cell swap, grafting a
fixed donor hull correlated $0.9$ with the label, so hull appearance is the spurious attribute and
the seam is identical on both sides. The \emph{hull-swap flip rate} tracks worst-group accuracy at
pooled $r{=}{-}0.941$ ($45$ configs; Fig.~\ref{fig:validation}b), ERM leaning on the graft and DFR
removing it. The correlation survives the nested design (cluster bootstrap over training runs,
$95\%$ CI $[-0.964,-0.916]$; leave-one-shortcut-out and Spearman both $\le{-}0.93$;
App.~\ref{app:circularity}). A swap to a third, never-seen donor hull falls in step (ERM $0.23$ to
DFR $0.12$; $r{=}0.73$ with the graft measure), so the operator reads hull reliance, not memory of
the planted donor. Where the planted strength is graded (cars, Sec.~\ref{sec:cars}), ERM's measured
reliance follows it ($0.82/0.30/0.05$ for strong/medium/weak).

This establishes that HullProbe orders and resolves known reliance correctly, not that it is a
ground truth on natural data, where cues are messier than planted ones. The measure is
interventional by construction, reading the decision change under a part replacement; agreement
with worst-group accuracy is convergent evidence for it, not its definition.

\begin{table}[t]\centering
\caption{\textbf{Replication across six planted shortcuts} (Sec.~\ref{sec:phase2}). Interventional
reliance (flip rate) for the untreated ERM model and the best de-biaser (DFR), and the
correlation between worst-group accuracy and reliance across all de-biaser$\times$seed configurations.
\emph{Appearance} shortcuts validate the measurement principle;
\emph{anatomical} shortcuts validate the swap operator itself (plant and measure both use the real
$6$-cell hull graft, listed by the two donor hulls). Watermark$^{\dagger}$ is a known null (no
shortcut learned $\Rightarrow$ near-zero measured reliance; its correlation, taken over that
near-constant range, is not informative).}
\label{tab:replication}
\resizebox{\columnwidth}{!}{%
\begin{tabular}{lccc}
\toprule
shortcut & ERM\,$\downarrow$ & DFR\,$\downarrow$ & corr(WGA, rel.) \\
\midrule
\multicolumn{4}{l}{\emph{appearance recolor} (validates the principle)}\\
\quad hull color  & $0.39$ & $\mathbf{0.12}$ & $-0.966$ \\
\quad background   & $0.43$ & $\mathbf{0.14}$ & $-0.904$ \\
\quad watermark$^{\dagger}$ & $0.08$ & $\mathbf{0.02}$ & $-0.891$ \\
\midrule
\multicolumn{4}{l}{\emph{anatomical part-swap} (validates the operator)}\\
\quad Veh-Carrier\,/\,Ro-ro     & $0.58$ & $\mathbf{0.09}$ & $-0.945$ \\
\quad Container\,/\,Wood-Chips  & $0.59$ & $\mathbf{0.15}$ & $-0.964$ \\
\quad Fishing\,/\,Yacht         & $0.52$ & $\mathbf{0.13}$ & $-0.928$ \\
\bottomrule
\end{tabular}}
\end{table}

\subsection{Robustness to segmentation, donors, and artifacts}
\label{sec:sensitivity}
\noindent\textbf{Segmentation.} The learned component is the hull/superstructure boundary, so we
repeat the audit on the $202$ annotated vessels the segmenter never trained on, with its predicted
maps and with the annotations, holding recipients, donors (drawn from the same set), silhouette, and
classifier fixed (App.~\ref{app:gtmask}). The segmenter disagrees with the annotation on $7\%$ of
vessel pixels (hull/superstructure IoU $0.86/0.85$), yet swapping in the annotations changes the
corrected hull-share by only $-0.009$ $[-0.041,0.023]$ ($n{=}194$; $0.49$ vs.\ $0.48$), every
per-cell influence by under $0.01$, and the spectrum not at all. These vessels are mostly from
MARVEL's training split, so we compare the arms, not the level: the segmenter's actual errors move
the hull-share by less than the audit's ${\sim}0.04$ resolution. The $\pm3$px test
(App.~Table~\ref{tab:sensitivity}) is a harsher, systematic perturbation that shifts every boundary
one way, unlike the nearly unbiased segmenter (superstructure fraction $0.48$ vs.\ $0.49$
annotated); it moves the hull-share by $-0.05$ (dilation) and $+0.10$ (erosion, which also strips
thin masts).

\smallskip\noindent\textbf{Donors and artifacts.} On a fixed $61$-recipient subsample, donor count and
selection leave the corrected hull-share at $0.50$--$0.52$ ($K\in\{2,4,8\}$, compatible or random
donors). The random-donor row also tests the artifact correction: shape-mismatched donors make worse
composites and raise the same-class floor, yet leave the corrected value unchanged, as the
class-independence assumption predicts.

\section{What HullProbe Reveals}
\label{sec:results}
We now apply the instrument to the natural MARVEL task: where the ConvNeXt-Tiny's sensitivity
lives (Secs.~\ref{sec:probe}--\ref{sec:spectrum}), whether it sets the class
(Sec.~\ref{sec:hijack}), whether de-biasing changes it (Sec.~\ref{sec:debias}), and whether the
picture generalizes (Sec.~\ref{sec:generality}). Brackets are $95\%$ bootstrap intervals.

\subsection{Hull vs.\ superstructure}
\label{sec:probe}
Swapping a full component ($400$ pairs) shows the average roles. Replacing the superstructure moves
the correct-class probability by $\Delta{=}0.251$ $[0.201,0.303]$, while a donor's hull transfers its
class only weakly ($\Delta{=}0.083$ $[0.053,0.114]$); keeping the superstructure retains the class
$66\%$ of the time vs.\ $39\%$ for the hull alone. The exhaustive cube ($n{=}277$) refines this.
Per-cell influence (App.~Table~\ref{tab:cells}) peaks at superstructure-mid but is non-trivial
everywhere, and summed by component the artifact-corrected sensitivity is $54\%$ superstructure and
$46\%$ hull (hull-share $0.46$ $[0.41,0.52]$). We read this as a directional split, not a precise
allocation, given the interval. The same-class control matters:
raw cross-class influence (mean $0.18$ over cells) is three times the corrected value ($0.06$), so
about two-thirds of naive ``part importance'' is generic within-class appearance change and
compositing.

\begin{figure}[t]
\centering
\begin{minipage}{0.49\linewidth}\centering
\includegraphics[width=\linewidth]{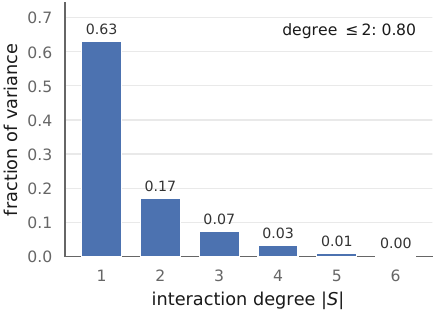}\\{\footnotesize(a) interaction degree}
\end{minipage}\hfill
\begin{minipage}{0.49\linewidth}\centering
\includegraphics[width=\linewidth]{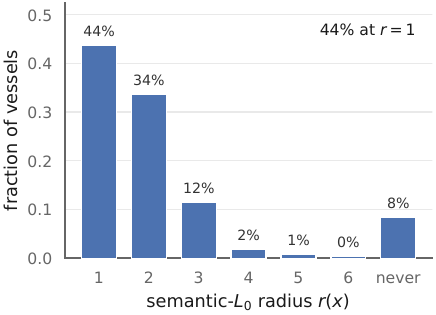}\\{\footnotesize(b) semantic-$L_0$}
\end{minipage}
\caption{\textbf{Nearly additive anatomy and single-swap fragility.} \emph{(a)} Variance by
interaction degree: $W_1{=}0.63$ of behavior is single-cell main effects and $0.80$ is degree
$\le2$, so per-cell attribution captures most of the decision (per-cell influence in
App.~Table~\ref{tab:cells}). \emph{(b)} Semantic-$L_0$ distribution: the fewest realistic part
swaps that flip the class; $44\%$ of vessels flip on one swap ($r{=}1$), the rest need
$\ge2$, and $8\%$ never flip within the six cells.}
\label{fig:spectrum}
\label{fig:semL0}
\end{figure}

\subsection{Nearly additive anatomy and single-swap fragility}
\label{sec:spectrum}
\label{sec:l0}
The interaction spectrum (Fig.~\ref{fig:spectrum}a) is concentrated at low order: main effects
hold $W_1{=}0.63$ $[0.60,0.66]$ of the configuration variance and $W_{\le2}{=}0.80$ $[0.77,0.83]$. The
donor-averaged response $g$ is therefore largely additive over cells, with secondary pairwise
interactions, so a per-cell reading captures most of its structure (a property of $g$, not of the
classifier's internal representation). The worst case is sharper. The semantic-$L_0$ radius
(Eq.~\eqref{eq:l0}; Fig.~\ref{fig:semL0}b) is $r{=}1$ for $44\%$ $[38,50]\%$ of vessels, where one
realistic cell swap flips the class; the median is $r{=}2$, and $8\%$ never flip. Because $r$ is a
minimum over a finite donor bank, $\Pr[r{=}1]$ grows with $K$ ($45\%\to52\%$ from $K{=}4$ to $8$ on
a common pool), so we treat it as ordinal (App.~\ref{app:robustness}).

\subsection{Sensitivity is not sufficiency}
\label{sec:hijack}
High hull sensitivity need not mean the hull is \emph{sufficient}~\cite{watson_ns} to set the class,
nor that relying on it is spurious; for types the hull genuinely defines, substantial reliance may
be appropriate. Under a direct conflict (a cross-class donor hull grafted on while the
superstructure still says the true class; $476$ conflicts, App.~\ref{app:sufficiency},
Fig.~\ref{fig:conflict}), the classifier follows the superstructure $73\%$ of the time and the donor
hull only $4.8\%$ (the rest default to a third class). The $46\%$ sensitivity thus shows up mostly
as degraded confidence, not a class-specific error: on this task the hull rarely overrides the class
it is grafted against.

\subsection{Auditing de-biasing without group labels}
\label{sec:debias}
\begin{table}[t]
\centering
\caption{\textbf{De-biasing audit} (3 seeds, mean$\pm$std). Hull-share is the balance
axis; $\Delta$ is the paired change vs.\ ERM (positive $=$ more hull-reliant);
$\Pr[r{=}1]$ is the concentration axis. No method reduces hull-share; CutMix trades
concentration for unchanged balance; bg-swap worsens both. $^{*}${:}\ $|\Delta|>2\sigma$.
Each $\Delta$ is averaged per seed over the recipients both that method and ERM classify
correctly, so it is a tighter paired estimate and need not equal the difference of the two
pooled hull-share columns (which use each method's own recipient pool).}
\label{tab:debias}
\resizebox{\columnwidth}{!}{%
\begin{tabular}{lccc}
\toprule
method & hull-share & $\Delta$ vs ERM & $\Pr[r{=}1]$ \\
\midrule
ERM             & $0.394{\scriptstyle\pm.003}$ & --- & $0.513$ \\
CutMix          & $0.396{\scriptstyle\pm.011}$ & $+0.002{\scriptstyle\pm.018}$ & $\mathbf{0.420}$ \\
hull-rand.      & $0.408{\scriptstyle\pm.011}$ & $+0.012{\scriptstyle\pm.008}$ & $0.548$ \\
JTT             & $0.399{\scriptstyle\pm.024}$ & $+0.002{\scriptstyle\pm.019}$ & $0.505$ \\
LfF             & $0.417{\scriptstyle\pm.019}$ & $+0.022{\scriptstyle\pm.019}$ & $0.538$ \\
AFR             & $0.399{\scriptstyle\pm.011}$ & $+0.006{\scriptstyle\pm.007}$ & $0.478$ \\
bg-swap         & $\mathbf{0.461}{\scriptstyle\pm.010}$ & $\mathbf{+0.069}{\scriptstyle\pm.010}^{*}$ & $0.733$ \\
\bottomrule
\end{tabular}}
\end{table}
We measure what de-biasers do to part reliance (Table~\ref{tab:debias}; App.~Fig.~\ref{fig:debias})
across three seeds, scoring each model on the recipients it classifies correctly (${\sim}70$ per
method) and pairing changes
on recipients both it and ERM get right (App.~\ref{app:debias}). Two axes move independently: the
hull-share (which parts carry the decision) and $\Pr[r{=}1]$ (how many must change to flip it). On
the first, no method lowers the hull-share. Every group-free paired change vs.\ ERM lies in
$[0,0.022]$, within seed noise (std ${\sim}0.02$), including hull-randomization, the augmentation
designed to break hull reliance, and AFR~\cite{afr}, the group-free counterpart of DFR. Only
background-swap moves significantly ($+0.069{\pm}0.010$). The audit's minimum detectable effect is
about $0.04$ (App.~\ref{app:debias}), so the null excludes reductions above that, not an exact
zero. It survives two checks of the ratio and of the correct-only population: absolute hull and
total influence are also flat, and a fixed-recipient audit with a continuous true-class response
leaves the soft hull-share equally flat (App.~Table~\ref{tab:absinf}).

The dividing line is group supervision. AFR fails here while DFR succeeds on planted shortcuts
(Sec.~\ref{sec:phase2}), because a method reduces hull-share only if it can target the shortcut.
Background-swap, the one method that moves, raises hull-share, consistent with forcing
background-invariance onto the vessel body. The number of parts moves separately: CutMix lowers
single-part fragility ($\Pr[r{=}1]$: $0.51\to0.42$) without making the model less hull-reliant,
background-swap raises it ($0.51\to0.73$), and the rest stay near ERM. Correlational attribution
barely registers any of this: Grad-CAM, Integrated Gradients, and RISE shift $\le0.05$ across
de-biasers where our interventional flip spans $0.27$ (App.~\ref{app:attribution}).

\subsection{Generality: architectures, datasets, and cars}
\label{sec:generality}
\noindent\textbf{Architectures.} The pooled corrected hull-share for ResNet-50~\cite{resnet} and
ViT-B/16~\cite{vit} baselines (same split; $75.4\%$/$73.9\%$ accuracy), each on its own correctly
classified recipients, is $0.46$ $[0.38,0.53]$ and $0.39$ $[0.31,0.47]$, against ConvNeXt-Tiny's
$0.48$ $[0.40,0.58]$ ($n{=}78/75/81$). The intervals overlap, so we read the shared level, not
the ordering: substantial hull reliance reflects the task rather than one backbone's inductive
bias.

\smallskip\noindent\textbf{A second maritime dataset.} On SeaShips~\cite{seaships} ($6$ types;
App.~\ref{app:seaships}), with the MARVEL segmenter applied zero-shot and a fresh classifier
($98.4\%$), the low-order structure strengthens ($W_{\le2}{=}0.91$ vs.\ $0.80$; hull-share $0.31$;
$25\%$ flip on one swap; $n{=}48$, indicative).

\smallskip\noindent\textbf{Beyond maritime: Stanford Cars.}\label{sec:cars} The vessel segmenter
and forward/mid/aft geometry do not transfer to cars~\cite{stanfordcars} ($196$ models), but the
protocol does once the part set is redefined: five parts (body, front-face, greenhouse, body-side,
wheel) from an open-vocabulary GroundedSAM~\cite{groundingdino,sam} prompt, registered by the same
PCA pose, with the same swaps, same-class control, and Walsh readouts on a $2^5$ cube
(App.~\ref{app:cars}). For a ConvNeXt-Tiny ($92.4\%$), recognition is again low-order
($W_1{=}0.94$, $W_{\le2}{=}0.99$; $95\%$ flip on one swap), and corrected influence concentrates on
the greenhouse (roof and glazing, $0.48$) far above the front-face ($0.04$), a non-obvious reading
recovered without car-specific tuning. Three graded color shortcuts (ResNet-50) reproduce the
convergent validity, recolor-flip reliance tracking worst-group accuracy at $r{=}{-}0.97$ over $45$
configurations (App.~Fig.~\ref{fig:carsconv}). The part-swap operator itself was validated only on
vessels, so the greenhouse reading rests on Sec.~\ref{sec:phase2}.

\section{Scope, Limitations, and Conclusion}
\label{sec:discussion}
\noindent\textbf{Where HullProbe applies.} A new domain must supply an exhaustive partition of the
object into a few named parts that every instance has, a closed-form frame registering each part
across instances, and a held-out pool with several instances per class to serve as donors. The swap
operator, artifact control, cube, and readouts carry over unchanged, and the classifier is queried
as a black box without group labels (App.~Table~\ref{tab:porting} lists every component). Vessels
and side-view cars meet these requirements, and other rigid, elongated objects seen from a
consistent viewpoint plausibly do. Articulated, deformable, or viewpoint-varying categories
(animals, people, most ImageNet classes) do not without learned correspondence, and the exhaustive
cube stays practical only up to about eight parts. HullProbe is thus a specialized instrument, not a
turnkey one, and our evidence for generality is empirical (three architectures, two maritime
datasets, one non-maritime domain): it shows that the protocol transfers, not that its findings will.

\smallskip\noindent\textbf{Limitations.} The estimate leans on one assumption, that same- and cross-class swaps carry comparable artifacts. The
random-donor test supports it but cannot rule out a cross-class incompatibility that mimics a class
cue; a mask-constrained generative edit, with a different artifact profile, is the natural
independent check. The composites are anatomically plausible but not photorealistic, the vessel
silhouette is shared with the annotations and so untested by Sec.~\ref{sec:sensitivity}, and the
longitudinal cells are pose-fixed canonical thirds, not a detected bow and stern. We therefore
trust the direction of a comparison but not a $0.03$--$0.05$ gap.

\smallskip\noindent\textbf{Release.} The evaluation harness, segmenter and annotations, composition
code, and the recipient and donor lists behind every reported number are at \releaseurl.

\smallskip\noindent\textbf{Conclusion.} HullProbe turns two orthogonal part decompositions into a
group-label-free measure of interventional part reliance that recovers planted shortcuts, is
unmoved when annotations replace its segmenter, and carries over to cars. Whether de-biasing has
removed a shortcut then depends on the method, which HullProbe reads directly.

{
    \small
    \bibliographystyle{ieeenat_fullname}
    \bibliography{refs}

\begin{thebibliography}{72}
\providecommand{\natexlab}[1]{#1}
\providecommand{\url}[1]{\texttt{#1}}
\expandafter\ifx\csname urlstyle\endcsname\relax
  \providecommand{\doi}[1]{doi: #1}\else
  \providecommand{\doi}{doi: \begingroup \urlstyle{rm}\Url}\fi

\bibitem[Adebayo et~al.(2018)Adebayo, Gilmer, Muelly, Goodfellow, Hardt, and
  Kim]{sanity}
Julius Adebayo, Justin Gilmer, Michael Muelly, Ian Goodfellow, Moritz Hardt,
  and Been Kim.
\newblock Sanity checks for saliency maps.
\newblock In \emph{NeurIPS}, 2018.

\bibitem[Arjovsky et~al.(2019)Arjovsky, Bottou, Gulrajani, and Lopez-Paz]{irm}
Martin Arjovsky, L{\'e}on Bottou, Ishaan Gulrajani, and David Lopez-Paz.
\newblock Invariant risk minimization.
\newblock \emph{arXiv:1907.02893}, 2019.

\bibitem[Bau et~al.(2017)Bau, Zhou, Khosla, Oliva, and Torralba]{netdissect}
David Bau, Bolei Zhou, Aditya Khosla, Aude Oliva, and Antonio Torralba.
\newblock Network dissection: Quantifying interpretability of deep visual
  representations.
\newblock In \emph{CVPR}, 2017.

\bibitem[Beery et~al.(2018)Beery, Van~Horn, and Perona]{terra}
Sara Beery, Grant Van~Horn, and Pietro Perona.
\newblock Recognition in terra incognita.
\newblock In \emph{ECCV}, 2018.

\bibitem[Brown et~al.(2017)Brown, Man{\'e}, Roy, Abadi, and Gilmer]{advpatch}
Tom~B Brown, Dandelion Man{\'e}, Aurko Roy, Mart{\'i}n Abadi, and Justin
  Gilmer.
\newblock Adversarial patch.
\newblock \emph{arXiv:1712.09665}, 2017.

\bibitem[Chang et~al.(2019)Chang, Creager, Goldenberg, and Duvenaud]{chang_cf}
Chun-Hao Chang, Elliot Creager, Anna Goldenberg, and David Duvenaud.
\newblock Explaining image classifiers by counterfactual generation.
\newblock In \emph{ICLR}, 2019.

\bibitem[Chen et~al.(2019)Chen, Li, Tao, Barnett, Rudin, and Su]{protopnet}
Chaofan Chen, Oscar Li, Daniel Tao, Alina Barnett, Cynthia Rudin, and
  Jonathan~K Su.
\newblock This looks like that: Deep learning for interpretable image
  recognition.
\newblock In \emph{NeurIPS}, 2019.

\bibitem[Covert et~al.(2020)Covert, Lundberg, and Lee]{sage}
Ian Covert, Scott Lundberg, and Su-In Lee.
\newblock Understanding global feature contributions with additive importance
  measures.
\newblock In \emph{NeurIPS}, 2020.

\bibitem[Creager et~al.(2021)Creager, Jacobsen, and Zemel]{eiil}
Elliot Creager, J{\"o}rn-Henrik Jacobsen, and Richard Zemel.
\newblock Environment inference for invariant learning.
\newblock In \emph{ICML}, 2021.

\bibitem[Cubuk et~al.(2020)Cubuk, Zoph, Shlens, and Le]{randaugment}
Ekin~D Cubuk, Barret Zoph, Jonathon Shlens, and Quoc~V Le.
\newblock {RandAugment}: Practical automated data augmentation with a reduced
  search space.
\newblock In \emph{NeurIPS}, 2020.

\bibitem[DeVries and Taylor(2017)]{cutout}
Terrance DeVries and Graham~W Taylor.
\newblock Improved regularization of convolutional neural networks with cutout.
\newblock \emph{arXiv:1708.04552}, 2017.

\bibitem[Dosovitskiy et~al.(2021)Dosovitskiy, Beyer, Kolesnikov, et~al.]{vit}
Alexey Dosovitskiy, Lucas Beyer, Alexander Kolesnikov, et~al.
\newblock An image is worth 16x16 words: Transformers for image recognition at
  scale.
\newblock In \emph{ICLR}, 2021.

\bibitem[Efron and Stein(1981)]{efron_stein}
Bradley Efron and Charles Stein.
\newblock The jackknife estimate of variance.
\newblock \emph{The Annals of Statistics}, 9\penalty0 (3):\penalty0 586--596,
  1981.

\bibitem[Fong and Vedaldi(2017)]{meaningful_pert}
Ruth~C Fong and Andrea Vedaldi.
\newblock Interpretable explanations of black boxes by meaningful perturbation.
\newblock In \emph{ICCV}, 2017.

\bibitem[Geirhos et~al.(2019)Geirhos, Rubisch, Michaelis, Bethge, Wichmann, and
  Brendel]{texturebias}
Robert Geirhos, Patricia Rubisch, Claudio Michaelis, Matthias Bethge, Felix~A
  Wichmann, and Wieland Brendel.
\newblock Imagenet-trained {CNN}s are biased towards texture; increasing shape
  bias improves accuracy and robustness.
\newblock In \emph{ICLR}, 2019.

\bibitem[Geirhos et~al.(2020)Geirhos, Jacobsen, Michaelis, Zemel, Brendel,
  Bethge, and Wichmann]{shortcut}
Robert Geirhos, J{\"o}rn-Henrik Jacobsen, Claudio Michaelis, Richard Zemel,
  Wieland Brendel, Matthias Bethge, and Felix~A Wichmann.
\newblock Shortcut learning in deep neural networks.
\newblock \emph{Nature Machine Intelligence}, 2020.

\bibitem[Ghorbani et~al.(2019)Ghorbani, Abid, and Zou]{fragile}
Amirata Ghorbani, Abubakar Abid, and James Zou.
\newblock Interpretation of neural networks is fragile.
\newblock In \emph{AAAI}, 2019.

\bibitem[Goodfellow et~al.(2015)Goodfellow, Shlens, and Szegedy]{fgsm}
Ian~J Goodfellow, Jonathon Shlens, and Christian Szegedy.
\newblock Explaining and harnessing adversarial examples.
\newblock In \emph{ICLR}, 2015.

\bibitem[Goyal et~al.(2019)Goyal, Wu, Ernst, Batra, Parikh, and Lee]{goyal_cf}
Yash Goyal, Ziyan Wu, Jan Ernst, Dhruv Batra, Devi Parikh, and Stefan Lee.
\newblock Counterfactual visual explanations.
\newblock In \emph{ICML}, 2019.

\bibitem[Gundogdu et~al.(2016)Gundogdu, Solmaz, Y{\"u}cesoy, and
  Ko{\c{c}}]{marvel}
Erhan Gundogdu, Berkan Solmaz, Veysel Y{\"u}cesoy, and Aykut Ko{\c{c}}.
\newblock {MARVEL}: A large-scale image dataset for maritime vessels.
\newblock In \emph{Asian Conference on Computer Vision (ACCV)}, 2016.

\bibitem[He et~al.(2016)He, Zhang, Ren, and Sun]{resnet}
Kaiming He, Xiangyu Zhang, Shaoqing Ren, and Jian Sun.
\newblock Deep residual learning for image recognition.
\newblock In \emph{CVPR}, 2016.

\bibitem[He et~al.(2017)He, Gkioxari, Doll{\'a}r, and Girshick]{maskrcnn}
Kaiming He, Georgia Gkioxari, Piotr Doll{\'a}r, and Ross Girshick.
\newblock Mask {R-CNN}.
\newblock In \emph{ICCV}, 2017.

\bibitem[Hendrycks et~al.(2020)Hendrycks, Mu, Cubuk, Zoph, Gilmer, and
  Lakshminarayanan]{augmix}
Dan Hendrycks, Norman Mu, Ekin~D Cubuk, Barret Zoph, Justin Gilmer, and Balaji
  Lakshminarayanan.
\newblock {AugMix}: A simple data processing method to improve robustness and
  uncertainty.
\newblock In \emph{ICLR}, 2020.

\bibitem[Hooker et~al.(2019)Hooker, Erhan, Kindermans, and Kim]{roar}
Sara Hooker, Dumitru Erhan, Pieter-Jan Kindermans, and Been Kim.
\newblock A benchmark for interpretability methods in deep neural networks.
\newblock In \emph{NeurIPS}, 2019.

\bibitem[Janzing et~al.(2020)Janzing, Minorics, and Bl{\"o}baum]{janzing}
Dominik Janzing, Lenon Minorics, and Patrick Bl{\"o}baum.
\newblock Feature relevance quantification in explainable {AI}: A causal
  problem.
\newblock In \emph{AISTATS}, 2020.

\bibitem[Kim et~al.(2018)Kim, Wattenberg, Gilmer, Cai, Wexler, Vi{\'e}gas, and
  Sayres]{tcav}
Been Kim, Martin Wattenberg, Justin Gilmer, Carrie Cai, James Wexler, Fernanda
  Vi{\'e}gas, and Rory Sayres.
\newblock Interpretability beyond feature attribution: Quantitative testing
  with concept activation vectors ({TCAV}).
\newblock In \emph{ICML}, 2018.

\bibitem[Kirichenko et~al.(2023)Kirichenko, Izmailov, and Wilson]{dfr}
Polina Kirichenko, Pavel Izmailov, and Andrew~Gordon Wilson.
\newblock Last layer re-training is sufficient for robustness to spurious
  correlations.
\newblock In \emph{ICLR}, 2023.

\bibitem[Kirillov et~al.(2023)Kirillov, Mintun, Ravi, Mao, Rolland, Gustafson,
  Xiao, Whitehead, Berg, Lo, Doll{\'a}r, and Girshick]{sam}
Alexander Kirillov, Eric Mintun, Nikhila Ravi, Hanzi Mao, Chloe Rolland, Laura
  Gustafson, Tete Xiao, Spencer Whitehead, Alexander~C Berg, Wan-Yen Lo, Piotr
  Doll{\'a}r, and Ross Girshick.
\newblock Segment anything.
\newblock In \emph{ICCV}, 2023.

\bibitem[Koh et~al.(2021)Koh, Sagawa, Marklund, et~al.]{wilds}
Pang~Wei Koh, Shiori Sagawa, Henrik Marklund, et~al.
\newblock {WILDS}: A benchmark of in-the-wild distribution shifts.
\newblock In \emph{ICML}, 2021.

\bibitem[Krause et~al.(2013)Krause, Stark, Deng, and Fei-Fei]{stanfordcars}
Jonathan Krause, Michael Stark, Jia Deng, and Li Fei-Fei.
\newblock 3d object representations for fine-grained categorization.
\newblock In \emph{IEEE International Conference on Computer Vision Workshops
  (3dRR)}, 2013.

\bibitem[Linial et~al.(1993)Linial, Mansour, and Nisan]{lmn}
Nathan Linial, Yishay Mansour, and Noam Nisan.
\newblock Constant depth circuits, fourier transform, and learnability.
\newblock \emph{Journal of the ACM}, 40\penalty0 (3):\penalty0 607--620, 1993.

\bibitem[Liu et~al.(2021)Liu, Haghgoo, Chen, Raghunathan, Koh, Sagawa, Liang,
  and Finn]{jtt}
Evan~Z Liu, Behzad Haghgoo, Annie~S Chen, Aditi Raghunathan, Pang~Wei Koh,
  Shiori Sagawa, Percy Liang, and Chelsea Finn.
\newblock Just train twice: Improving group robustness without training group
  information.
\newblock In \emph{ICML}, 2021.

\bibitem[Liu et~al.(2024)Liu, Zeng, Ren, Li, Zhang, Yang, Jiang, Li, Yang, Su,
  Zhu, and Zhang]{groundingdino}
Shilong Liu, Zhaoyang Zeng, Tianhe Ren, Feng Li, Hao Zhang, Jie Yang, Qing
  Jiang, Chunyuan Li, Jianwei Yang, Hang Su, Jun Zhu, and Lei Zhang.
\newblock Grounding {DINO}: Marrying {DINO} with grounded pre-training for
  open-set object detection.
\newblock In \emph{ECCV}, 2024.

\bibitem[Liu et~al.(2022)Liu, Mao, Wu, Feichtenhofer, Darrell, and
  Xie]{convnext}
Zhuang Liu, Hanzi Mao, Chao-Yuan Wu, Christoph Feichtenhofer, Trevor Darrell,
  and Saining Xie.
\newblock A convnet for the 2020s.
\newblock In \emph{CVPR}, 2022.

\bibitem[Lundberg and Lee(2017)]{shap}
Scott~M Lundberg and Su-In Lee.
\newblock A unified approach to interpreting model predictions.
\newblock In \emph{NeurIPS}, 2017.

\bibitem[Madry et~al.(2018)Madry, Makelov, Schmidt, Tsipras, and Vladu]{pgd}
Aleksander Madry, Aleksandar Makelov, Ludwig Schmidt, Dimitris Tsipras, and
  Adrian Vladu.
\newblock Towards deep learning models resistant to adversarial attacks.
\newblock In \emph{ICLR}, 2018.

\bibitem[Nam et~al.(2020)Nam, Cha, Ahn, Lee, and Shin]{lff}
Junhyun Nam, Hyuntak Cha, Sungsoo Ahn, Jaeho Lee, and Jinwoo Shin.
\newblock Learning from failure: De-biasing classifier from biased classifier.
\newblock In \emph{NeurIPS}, 2020.

\bibitem[O'Donnell(2014)]{odonnell}
Ryan O'Donnell.
\newblock \emph{Analysis of Boolean Functions}.
\newblock Cambridge University Press, 2014.

\bibitem[Owen(2014)]{owen_shapley}
Art~B Owen.
\newblock Sobol' indices and shapley value.
\newblock \emph{SIAM/ASA Journal on Uncertainty Quantification}, 2014.

\bibitem[Papernot et~al.(2016)Papernot, McDaniel, Jha, Fredrikson, Celik, and
  Swami]{jsma}
Nicolas Papernot, Patrick McDaniel, Somesh Jha, Matt Fredrikson, Z~Berkay
  Celik, and Ananthram Swami.
\newblock The limitations of deep learning in adversarial settings.
\newblock In \emph{EuroS\&P}, 2016.

\bibitem[Pearl(2009)]{pearl}
Judea Pearl.
\newblock \emph{Causality: Models, Reasoning, and Inference}.
\newblock Cambridge University Press, 2009.

\bibitem[Petsiuk et~al.(2018)Petsiuk, Das, and Saenko]{rise}
Vitali Petsiuk, Abir Das, and Kate Saenko.
\newblock {RISE}: Randomized input sampling for explanation of black-box
  models.
\newblock In \emph{BMVC}, 2018.

\bibitem[Prasad et~al.(2017)Prasad, Rajan, Rachmawati, Rajabally, and
  Quek]{smd}
Dilip~K Prasad, Deepu Rajan, Lily Rachmawati, Eshan Rajabally, and Chai Quek.
\newblock Video processing from electro-optical sensors for object detection
  and tracking in a maritime environment: A survey.
\newblock \emph{IEEE Transactions on Intelligent Transportation Systems}, 2017.

\bibitem[Qiu et~al.(2023)Qiu, Potapczynski, Izmailov, and Wilson]{afr}
Shikai Qiu, Andres Potapczynski, Pavel Izmailov, and Andrew~Gordon Wilson.
\newblock Simple and fast group robustness by automatic feature reweighting.
\newblock In \emph{ICML}, 2023.

\bibitem[Ribeiro et~al.(2016)Ribeiro, Singh, and Guestrin]{lime}
Marco~Tulio Ribeiro, Sameer Singh, and Carlos Guestrin.
\newblock ``why should i trust you?'': Explaining the predictions of any
  classifier.
\newblock In \emph{KDD}, 2016.

\bibitem[Ross et~al.(2017)Ross, Hughes, and Doshi-Velez]{ross_rrr}
Andrew~Slavin Ross, Michael~C Hughes, and Finale Doshi-Velez.
\newblock Right for the right reasons: Training differentiable models by
  constraining their explanations.
\newblock In \emph{IJCAI}, 2017.

\bibitem[Sagawa et~al.(2020)Sagawa, Koh, Hashimoto, and Liang]{groupdro}
Shiori Sagawa, Pang~Wei Koh, Tatsunori~B Hashimoto, and Percy Liang.
\newblock Distributionally robust neural networks for group shifts: On the
  importance of regularization for worst-case generalization.
\newblock In \emph{ICLR}, 2020.

\bibitem[Sauer and Geiger(2021)]{sauer_cgn}
Axel Sauer and Andreas Geiger.
\newblock Counterfactual generative networks.
\newblock In \emph{ICLR}, 2021.

\bibitem[Sch{\"o}lkopf et~al.(2021)Sch{\"o}lkopf, Locatello, Bauer, Ke,
  Kalchbrenner, Goyal, and Bengio]{causal_rep}
Bernhard Sch{\"o}lkopf, Francesco Locatello, Stefan Bauer, Nan~Rosemary Ke, Nal
  Kalchbrenner, Anirudh Goyal, and Yoshua Bengio.
\newblock Toward causal representation learning.
\newblock \emph{Proceedings of the IEEE}, 2021.

\bibitem[Selvaraju et~al.(2017)Selvaraju, Cogswell, Das, Vedantam, Parikh, and
  Batra]{gradcam}
Ramprasaath~R Selvaraju, Michael Cogswell, Abhishek Das, Ramakrishna Vedantam,
  Devi Parikh, and Dhruv Batra.
\newblock {Grad-CAM}: Visual explanations from deep networks via gradient-based
  localization.
\newblock In \emph{ICCV}, 2017.

\bibitem[Shao et~al.(2018)Shao, Wu, Wang, Du, and Li]{seaships}
Zhenfeng Shao, Wenjing Wu, Zhongyuan Wang, Wan Du, and Chengyuan Li.
\newblock {SeaShips}: A large-scale precisely annotated dataset for ship
  detection.
\newblock \emph{IEEE Transactions on Multimedia}, 20\penalty0 (10):\penalty0
  2593--2604, 2018.

\bibitem[Shrikumar et~al.(2017)Shrikumar, Greenside, and Kundaje]{deeplift}
Avanti Shrikumar, Peyton Greenside, and Anshul Kundaje.
\newblock Learning important features through propagating activation
  differences.
\newblock In \emph{ICML}, 2017.

\bibitem[Singla and Feizi(2022)]{salientimagenet}
Sahil Singla and Soheil Feizi.
\newblock Salient {ImageNet}: How to discover spurious features in deep
  learning?
\newblock In \emph{ICLR}, 2022.

\bibitem[Smilkov et~al.(2017)Smilkov, Thorat, Kim, Vi{\'e}gas, and
  Wattenberg]{smoothgrad}
Daniel Smilkov, Nikhil Thorat, Been Kim, Fernanda Vi{\'e}gas, and Martin
  Wattenberg.
\newblock {SmoothGrad}: Removing noise by adding noise.
\newblock \emph{arXiv:1706.03825}, 2017.

\bibitem[Sobol(2001)]{sobol}
Ilya~M Sobol.
\newblock Global sensitivity indices for nonlinear mathematical models and
  their monte carlo estimates.
\newblock \emph{Mathematics and Computers in Simulation}, 55\penalty0
  (1-3):\penalty0 271--280, 2001.

\bibitem[Sohoni et~al.(2020)Sohoni, Dunnmon, Angus, Gu, and R{\'e}]{george}
Nimit Sohoni, Jared Dunnmon, Geoffrey Angus, Albert Gu, and Christopher R{\'e}.
\newblock No subclass left behind: Fine-grained robustness in coarse-grained
  classification problems.
\newblock In \emph{NeurIPS}, 2020.

\bibitem[Su et~al.(2019)Su, Vargas, and Sakurai]{onepixel}
Jiawei Su, Danilo~Vasconcellos Vargas, and Kouichi Sakurai.
\newblock One pixel attack for fooling deep neural networks.
\newblock \emph{IEEE Transactions on Evolutionary Computation}, 23\penalty0
  (5):\penalty0 828--841, 2019.

\bibitem[Sundararajan et~al.(2017)Sundararajan, Taly, and Yan]{ig}
Mukund Sundararajan, Ankur Taly, and Qiqi Yan.
\newblock Axiomatic attribution for deep networks.
\newblock In \emph{ICML}, 2017.

\bibitem[Taghanaki et~al.(2022)Taghanaki, Khani, Khani, Gholami, Tran,
  Mahdavi-Amiri, and Hamarneh]{masktune}
Saeid~Asgari Taghanaki, Aliasghar Khani, Fereshte Khani, Ali Gholami, Linh
  Tran, Ali Mahdavi-Amiri, and Ghassan Hamarneh.
\newblock {MaskTune}: Mitigating spurious correlations by forcing to explore.
\newblock In \emph{NeurIPS}, 2022.

\bibitem[Veitch et~al.(2021)Veitch, D'Amour, Yadlowsky, and Eisenstein]{veitch}
Victor Veitch, Alexander D'Amour, Steve Yadlowsky, and Jacob Eisenstein.
\newblock Counterfactual invariance to spurious correlations in text
  classification.
\newblock In \emph{NeurIPS}, 2021.

\bibitem[Wang et~al.(2021)Wang, Lyle, and Kwiatkowska]{interv_robust}
Benjie Wang, Clare Lyle, and Marta Kwiatkowska.
\newblock Provable guarantees on the robustness of decision rules to causal
  interventions.
\newblock In \emph{IJCAI}, 2021.

\bibitem[Watson et~al.(2021)Watson, Gultchin, Taly, and Floridi]{watson_ns}
David~S Watson, Limor Gultchin, Ankur Taly, and Luciano Floridi.
\newblock Local explanations via necessity and sufficiency: Unifying theory and
  practice.
\newblock In \emph{UAI}, 2021.

\bibitem[Xiao et~al.(2021)Xiao, Engstrom, Ilyas, and Madry]{backgrounds}
Kai Xiao, Logan Engstrom, Andrew Ilyas, and Aleksander Madry.
\newblock Noise or signal: The role of image backgrounds in object recognition.
\newblock In \emph{ICLR}, 2021.

\bibitem[Xie et~al.(2021)Xie, Wang, Yu, Anandkumar, Alvarez, and
  Luo]{segformer}
Enze Xie, Wenhai Wang, Zhiding Yu, Anima Anandkumar, Jose~M Alvarez, and Ping
  Luo.
\newblock {SegFormer}: Simple and efficient design for semantic segmentation
  with transformers.
\newblock In \emph{NeurIPS}, 2021.

\bibitem[Yang et~al.(2023)Yang, Zhang, Katabi, and Ghassemi]{subpop}
Yuzhe Yang, Haoran Zhang, Dina Katabi, and Marzyeh Ghassemi.
\newblock Change is hard: A closer look at subpopulation shift.
\newblock In \emph{ICML}, 2023.

\bibitem[Yao et~al.(2022)Yao, Wang, Li, Zhang, Liang, Zou, and Finn]{lisa}
Huaxiu Yao, Yu Wang, Sai Li, Linjun Zhang, Weixin Liang, James Zou, and Chelsea
  Finn.
\newblock Improving out-of-distribution robustness via selective augmentation.
\newblock In \emph{ICML}, 2022.

\bibitem[Youbi~Idrissi et~al.(2022)Youbi~Idrissi, Arjovsky, Pezeshki, and
  Lopez-Paz]{subg}
Badr Youbi~Idrissi, Martin Arjovsky, Mohammad Pezeshki, and David Lopez-Paz.
\newblock Simple data balancing achieves competitive worst-group-accuracy.
\newblock In \emph{CLeaR}, 2022.

\bibitem[Yun et~al.(2019)Yun, Han, Oh, Chun, Choe, and Yoo]{cutmix}
Sangdoo Yun, Dongyoon Han, Seong~Joon Oh, Sanghyuk Chun, Junsuk Choe, and
  Youngjoon Yoo.
\newblock {CutMix}: Regularization strategy to train strong classifiers with
  localizable features.
\newblock In \emph{ICCV}, 2019.

\bibitem[Zhang et~al.(2018)Zhang, Cisse, Dauphin, and Lopez-Paz]{mixup}
Hongyi Zhang, Moustapha Cisse, Yann~N Dauphin, and David Lopez-Paz.
\newblock mixup: Beyond empirical risk minimization.
\newblock In \emph{ICLR}, 2018.

\bibitem[Zhang et~al.(2022)Zhang, Sohoni, Zhang, Finn, and R{\'e}]{cnc}
Michael Zhang, Nimit~S Sohoni, Hongyang~R Zhang, Chelsea Finn, and Christopher
  R{\'e}.
\newblock Correct-n-contrast: A contrastive approach for improving robustness
  to spurious correlations.
\newblock In \emph{ICML}, 2022.

\bibitem[Zheng and Zhang(2020)]{mcships}
Yitong Zheng and Shun Zhang.
\newblock {McShips}: A large-scale ship dataset for detection and fine-grained
  categorization in the wild.
\newblock In \emph{ICME}, 2020.

\bibitem[Zhou et~al.(2017)Zhou, Zhao, Puig, Fidler, Barriuso, and
  Torralba]{ade20k}
Bolei Zhou, Hang Zhao, Xavier Puig, Sanja Fidler, Adela Barriuso, and Antonio
  Torralba.
\newblock Scene parsing through {ADE20K} dataset.
\newblock In \emph{CVPR}, 2017.

\end{thebibliography}
}
\clearpage
\setcounter{page}{1}
\maketitlesupplementary
\setcounter{section}{0}
\renewcommand{\thesection}{\Alph{section}}

\section{Segmentation error: annotations vs.\ the segmenter}
\label{app:gtmask}
Sec.~\ref{sec:sensitivity} asks how the segmenter's errors propagate into HullProbe. This appendix
gives the design and the full results (Table~\ref{tab:gtmask}).

\medskip\noindent\textbf{Design.} The segmenter's $80/20$ split of the $1{,}014$ annotations was drawn
with a fixed seed, so its $202$ validation vessels, never used in training, can be recovered exactly
(the per-class counts match the training log); they cover all $26$ types ($1$--$32$ each). We audit
them with the attribution classifier (ConvNeXt-Tiny) in three arms: with the segmenter's predicted
hull/superstructure maps for recipient and donors (\emph{segmenter}, the deployed pipeline), with the
annotations for both (\emph{annotation}), and with annotations for the recipient only, which
separates the two sides of the swap. Everything else is fixed across arms: the recipients, the donor
identities (drawn once: $K{=}4$ aspect-compatible cross-class donors and $2$ same-class controls, all
from the same held-out set, so every donor also has an annotation the segmenter never saw), the
vessel silhouette and PCA pose, and the classifier. Both map sources share the silhouette (the
zero-shot boat mask, App.~\ref{sec:operator-appendix}), so the arms differ only in where the
hull/superstructure boundary lies, which is the learned component and the one that decides the
hull-share. Of the $202$ vessels, $194$ are correctly classified and have a same-class control, and enter the audit. Only $22$ of
the $202$ are also in MARVEL's test split; the rest come from its training split, which the
classifier saw. That lowers absolute flip rates (a classifier is more robust on its own training
images) and the recipients are not class-balanced, so the absolute values here are not comparable
to the main pool's; the paired comparison is unaffected, since images, donors, and classifier are
identical across arms. We report the test-split vessels separately ($17$ of the $22$ enter the audit).

\medskip\noindent\textbf{What the annotations are.} Each annotation is a deck line placed by an
annotator, so its hull/superstructure boundary is straight, whereas the segmenter's follows the image
and can curve around deck fittings. Part of the disagreement below is therefore a difference in
boundary style rather than segmenter error, and the annotation arm is a reference, not an oracle.

\medskip\noindent\textbf{Results.} On these $194$ vessels the segmenter disagrees with the annotation
on $7.4\%$ of vessel pixels on average (median $5.3\%$; within-silhouette hull and superstructure IoU
$0.86$ and $0.85$), and its errors are nearly unbiased: it labels $48.0\%$ of vessel pixels
superstructure against the annotations' $48.8\%$. Replacing its maps with the annotations changes
the corrected hull-share by $-0.009$ $[-0.041,+0.023]$ (paired bootstrap over recipients), the raw
hull-share by $-0.015$, and the spectrum not at all ($W_1$ $+0.004$, $W_{\le2}$ $0.000$,
$\Pr[r{=}1]$ $0.000$). Every per-cell corrected influence moves by less than $0.01$ with an interval
covering zero, and the per-recipient corrected-influence profiles of the two arms correlate at
$r{=}0.79$, so the agreement holds recipient by recipient, not only on average. Annotating only the
recipient gives the same answer ($-0.013$ $[-0.044,+0.018]$), so donor-side segmentation error
contributes nothing measurable. The $17$ test-split vessels agree ($-0.002$), with an interval too
wide (${\pm}0.15$) to say more on their own.

Two conclusions follow. First, the segmenter's actual errors move the hull-share by less than the
audit's ${\sim}0.04$ resolution, so the hull-share is not a segmentation artifact. Second, the
$\pm3$px stress test of App.~\ref{app:robustness} overstates segmentation sensitivity. It shifts
every boundary in the same direction, a systematic bias the segmenter does not have, so even
dilation, which relabels fewer pixels ($5\%$) than the segmenter gets wrong ($7.4\%$), moves the
hull-share by $-0.05$. We keep those rows as a bracket for a segmenter that is biased toward one
class.

\begin{table*}[t]
\centering\small
\caption{\textbf{Segmentation error: annotations vs.\ the segmenter} (App.~\ref{app:gtmask}). The audit
on the segmenter's $202$ held-out vessels with its predicted hull/superstructure maps, with the
annotations, and with annotations for the recipient only; recipients, donors, silhouette, and
classifier are identical across arms. Brackets are $95\%$ recipient-bootstrap intervals; $\Delta$
rows are paired (annotation minus segmenter). \emph{Left:} pooled readouts. \emph{Right:}
per-cell corrected influence $\Inf_p^{\text{corr}}$ on all held-out vessels.}
\label{tab:gtmask}
\begin{minipage}{0.60\linewidth}\centering
\resizebox{\linewidth}{!}{%
\begin{tabular}{lccccc}
\toprule
mask source & hull-share & raw & $W_1$ & $W_{\le2}$ & $\Pr[r{=}1]$ \\
\midrule
\multicolumn{6}{l}{\emph{all held-out} ($n{=}194$)}\\
\quad segmenter (both)            & $0.49$ {\scriptsize$[0.45,0.53]$} & $0.47$ & $0.67$ & $0.84$ & $0.35$ \\
\quad annotation (both)           & $0.48$ {\scriptsize$[0.44,0.53]$} & $0.46$ & $0.67$ & $0.84$ & $0.35$ \\
\quad annotation (recipient only) & $0.48$ {\scriptsize$[0.44,0.52]$} & $0.46$ & $0.67$ & $0.84$ & $0.35$ \\
\quad $\Delta$ (both)             & $-0.009$ {\scriptsize$[-0.041,0.023]$} & $-0.015$ & $+0.004$ & $0.000$ & $0.000$ \\
\midrule
\multicolumn{6}{l}{\emph{held-out and in MARVEL's test split} ($n{=}17$)}\\
\quad segmenter (both)            & $0.48$ {\scriptsize$[0.34,0.61]$} & $0.45$ & $0.66$ & $0.85$ & $0.41$ \\
\quad annotation (both)           & $0.48$ {\scriptsize$[0.37,0.60]$} & $0.43$ & $0.67$ & $0.82$ & $0.41$ \\
\quad $\Delta$ (both)             & $-0.002$ {\scriptsize$[-0.140,0.152]$} & $-0.025$ & $+0.012$ & $-0.026$ & $0.000$ \\
\bottomrule
\end{tabular}}
\end{minipage}\hfill
\begin{minipage}{0.37\linewidth}\centering
\resizebox{\linewidth}{!}{%
\begin{tabular}{lccc}
\toprule
cell & segmenter & annotation & $\Delta$ \\
\midrule
hull-fwd & $0.075$ & $0.067$ & $-0.008$ {\scriptsize$[-0.019,0.002]$} \\
hull-mid & $0.132$ & $0.131$ & $-0.001$ {\scriptsize$[-0.014,0.011]$} \\
hull-aft & $0.068$ & $0.074$ & $+0.006$ {\scriptsize$[-0.005,0.016]$} \\
sup-fwd  & $0.057$ & $0.059$ & $+0.002$ {\scriptsize$[-0.009,0.014]$} \\
sup-mid  & $0.154$ & $0.162$ & $+0.009$ {\scriptsize$[-0.009,0.026]$} \\
sup-aft  & $0.074$ & $0.070$ & $-0.005$ {\scriptsize$[-0.018,0.009]$} \\
\bottomrule
\end{tabular}}
\end{minipage}
\end{table*}

\section{Robustness of the hull-share}
\label{app:robustness}
The corrected hull-share is stable under the choices in the audit (Table~\ref{tab:sensitivity}).
Donor selection and count are irrelevant (compatible vs.\ random donors, $K\in\{2,4,8\}$: all
$0.50$--$0.52$), and the random-donor row validates the artifact correction. Shape-mismatched
donors make worse composites yet leave the corrected value unchanged (raw $0.46$ vs.\ $0.47$;
corrected $0.52$ vs.\ $0.51$), so the same-class control
(Eq.~\eqref{eq:corr}) subtracts the compositing artifact, not the class signal.

\medskip\noindent\textbf{Boundary stress test.} Shifting the hull/superstructure boundary by $3$px
moves the value in the expected direction, by $-0.05$ when the superstructure is dilated and by
$+0.10$ when it is eroded, leaving the ordinal conclusion (substantial hull reliance) intact. The
two directions are not symmetric perturbations. Dilation only moves the internal boundary,
relabeling $5\%$ of vessel pixels on average over $400$ test-pool vessels, whereas erosion also
strips a $3$px band from the superstructure's outer edge, deleting thin masts and crane booms, and
relabels $15\%$. The perturbation is applied to the recipient's swap stencil; donor maps and the
per-cell fit are left unperturbed. App.~\ref{app:gtmask} replaces this synthetic perturbation with
the segmenter's actual error.

\medskip\noindent\textbf{Donor count and the fragility distribution.} The $K\in\{2,4,8\}$ check above
covers the pooled hull-share, which is a mean and is stable. The semantic-$L_0$ radius is different:
it is a \emph{min} over the $63$ non-empty subsets against a finite donor bank, so a larger bank can
only find more flipping swaps and drive the radius down. Recomputing it with $K{=}4$ and $K{=}8$
donors on a common $221$-recipient pool (same seed and recipients, a smaller draw than the
$277$-recipient headline pool) bears this out, raising the fraction that flips on one swap from
$45\%$ to $52\%$ (median $r$ tightening from $2$ to $1$). The radius is therefore
donor-count-sensitive, but in the opposite direction to the usual worry: $K{=}4$
\emph{under}-estimates fragility rather than inflating it, so the headline $44\%$ is a
conservative lower bound. The de-biasing audit's ERM baseline separately reads
$\Pr[r{=}1]{=}0.51$ on its own pool of correctly classified recipients (Table~\ref{tab:debias});
this $0.44$--$0.52$ spread across pool and $K$ is why we treat this fraction as pool- and
$K$-dependent. We accordingly treat the fragility
distribution (Fig.~\ref{fig:semL0}b) as ordinal and indicative, and lean on the pooled hull-share
and the interaction spectrum for the quantitative claims.

\begin{table}[t]\centering
\caption{\textbf{Sensitivity} of the corrected hull-share [$95\%$ bootstrap CI] to donor and
segmentation choices (App.~\ref{app:robustness}; fixed $61$-recipient subsample, which runs at
$0.50$--$0.52$, above the $0.46$ full-pool headline but within its bootstrap interval). Robust to donor
selection and count; the corrected value is invariant to donor quality (random vs.\ compatible),
validating the artifact control; segmentation boundary shifts move it in the expected direction.}
\label{tab:sensitivity}
\resizebox{\columnwidth}{!}{%
\begin{tabular}{lcc}
\toprule
variant & corrected hull-share & raw \\
\midrule
reference (compatible, $K{=}4$) & $0.51$ {\scriptsize$[.40,.63]$} & $0.47$ \\
random donors ($K{=}4$)         & $0.52$ {\scriptsize$[.40,.66]$} & $0.46$ \\
compatible, $K{=}2$             & $0.50$ {\scriptsize$[.35,.63]$} & $0.45$ \\
compatible, $K{=}8$             & $0.51$ {\scriptsize$[.42,.62]$} & $0.46$ \\
seg.\ eroded ($-3$px)           & $0.61$ {\scriptsize$[.51,.72]$} & $0.56$ \\
seg.\ dilated ($+3$px)          & $0.46$ {\scriptsize$[.36,.57]$} & $0.43$ \\
\bottomrule
\end{tabular}}
\end{table}

\section{Circularity control for the operator validation}
\label{app:circularity}
The operator-level validation of Sec.~\ref{sec:phase2} uses the same $6$-cell graft to plant the
shortcut (hull $\leftarrow$ donor $D_0$ or $D_1$) and to measure it (re-swapping the hull to the
other trained donor $D_{1-s}$). A high flip rate might then reflect nothing more than a memorized
$D_0$/$D_1$ hull, or leftover compositing statistics, instead of reliance that survives a change of donor. The
compositing worry is already handled, since the seam is identical across $s$ and cannot drive the flip.
Memorization is not. (The single $D_0/D_1$ shortcut used here is also learned almost completely,
untreated WGA${\approx}0.03$ against the ${\approx}0.20$ pooled over three shortcuts in
Table~\ref{tab:replication}, which is why its flip rates sit above the pooled numbers of
Table~\ref{tab:phase2detail}, e.g.\ ERM $0.79$ vs.\ $0.56$; only the ERM$\to$DFR drop and the
correlation carry over, not the absolute levels.)

To break the memorization route we swap the hull to a \emph{third} donor $D_2$, drawn from a class the
model never saw in training and composited the same way. A flip here can only come from a hull
appearance the model never learned. This independent-donor flip falls from $0.23$ at ERM to $0.12$ at
DFR, in step with the re-swap flip's $0.79\!\to\!0.08$, and the two agree across methods at mean
$r{=}0.73$ (Table~\ref{tab:circularity}, five de-biasers, three seeds). The reliance the operator
picks up therefore outlasts the two planted donors. It is a weaker and noisier signal than the
re-swap, as a never-seen hull should be, so the test loosens the plant-and-measure coupling without
severing it; only a physically re-hulled set would do that.

\medskip\noindent\textbf{Dependence in the pooled correlation.} The $45$ operator-level
configurations (Sec.~\ref{sec:phase2}) are not $45$ independent units, since they share three shortcut
constructions, one donor bank, five de-biasers, and three seeds, so a pooled Pearson $r$ can
overstate the evidence when the method ordering repeats across related conditions. We therefore
report the pooled $r{=}{-}0.941$ alongside the per-shortcut correlations ($-0.945$, $-0.964$,
$-0.928$; Table~\ref{tab:replication}), which are consistent across the three constructions.
Respecting the nested design confirms the pooled estimate rather than weakening it. A cluster
bootstrap over training runs ($9$ shortcut$\times$seed clusters, $20$k resamples) gives a $95\%$ CI of
$[-0.964,-0.916]$; a leave-one-out over shortcuts stays at $-0.946$/$-0.932$/$-0.950$ (each on the
other $30$ configurations); and the rank correlation is Spearman $\rho{=}{-}0.933$. The
association is thus not an artifact of pseudo-replication across the shared shortcuts and seeds
(Table~\ref{tab:robust}).

\begin{table}[t]\centering
\caption{\textbf{The operator-level correlation is robust to the nested design} ($45$ configs;
Sec.~\ref{sec:phase2}): reliance vs.\ worst-group accuracy stays strongly negative under every
check, so it is not an artifact of the shared shortcuts and seeds.}
\label{tab:robust}
\begin{tabular}{@{}lc@{}}
\toprule
robustness check & corr(WGA, reliance) \\
\midrule
pooled Pearson $r$          & $-0.941$ \\
Spearman $\rho$             & $-0.933$ \\
per-shortcut ($3$)          & $-0.945,\ -0.964,\ -0.928$ \\
leave-one-out (by shortcut)  & $-0.946,\ -0.932,\ -0.950$ \\
cluster bootstrap $95\%$ CI & $[-0.964,\ -0.916]$ \\
\bottomrule
\end{tabular}
\end{table}

\begin{table}[t]\centering
\caption{\textbf{Circularity control} (ResNet-50, $3$ seeds, mean$\pm$std; App.~\ref{app:circularity}).
For each de-biaser, the operator's re-swap flip (hull $\to$ the other \emph{trained} donor) and an
independent-mechanism flip (hull $\to$ a third, \emph{unseen} donor through the identical compositor).
Both fall from ERM to DFR and are positively correlated across methods (mean $r{=}0.73$), so the
operator reads hull reliance that survives a change of donor, not donor or compositing memorization.}
\label{tab:circularity}
\begin{tabular}{lccc}
\toprule
method & WGA & re-swap flip & indep.\ flip \\
\midrule
ERM      & $0.03{\scriptstyle\pm.03}$ & $0.79{\scriptstyle\pm.05}$ & $0.23{\scriptstyle\pm.05}$ \\
SUBG     & $0.05{\scriptstyle\pm.08}$ & $0.75{\scriptstyle\pm.12}$ & $0.21{\scriptstyle\pm.03}$ \\
GroupDRO & $0.21{\scriptstyle\pm.07}$ & $0.54{\scriptstyle\pm.14}$ & $0.18{\scriptstyle\pm.04}$ \\
JTT      & $0.13{\scriptstyle\pm.08}$ & $0.65{\scriptstyle\pm.09}$ & $0.17{\scriptstyle\pm.03}$ \\
DFR      & $\mathbf{0.65}{\scriptstyle\pm.07}$ & $\mathbf{0.08}{\scriptstyle\pm.03}$ & $\mathbf{0.12}{\scriptstyle\pm.06}$ \\
\bottomrule
\end{tabular}
\end{table}

\section{Hull sensitivity vs.\ sufficiency}
\label{app:sufficiency}
Section~\ref{sec:hijack} separates two questions that are easy to conflate. The first is how sensitive
the decision is to the hull (its influence share, $46\%$ globally); the second is whether the hull is
sufficient to set the predicted class. High sensitivity need not mean the classifier is
dominated by the hull, and the gap between the two is what shows the measured reliance is largely
benign on the natural task.

\begin{figure}[t]
\centering
\includegraphics[width=0.95\linewidth]{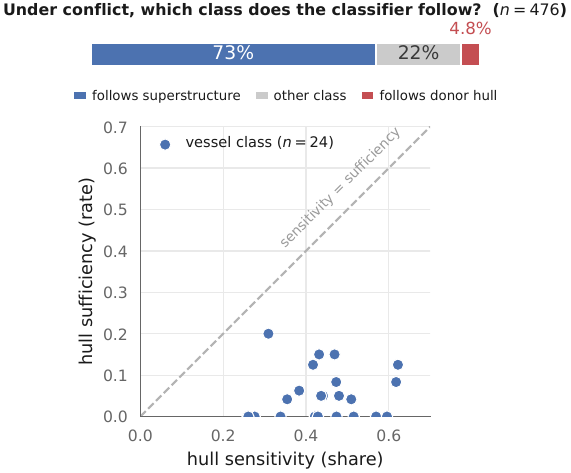}
\caption{\textbf{Hull sensitivity vs.\ sufficiency} (Sec.~\ref{sec:hijack}).
\emph{Top:} under a hull/superstructure conflict (a cross-class donor hull grafted on,
superstructure kept), the classifier follows the superstructure (the true class) $73\%$
of the time and adopts the donor hull's class only $4.8\%$. \emph{Bottom:} per class,
the hull sensitivity (share) far exceeds the rate at which the hull is actually
sufficient to set the prediction. Every class sits well below the $y{=}x$ line, so sensitivity
should not be read as harmful reliance.}
\label{fig:conflict}
\end{figure}

We probe sufficiency with a direct conflict. For each of $476$ recipient--donor pairs we graft a
cross-class donor hull onto the recipient while keeping its own superstructure, so that the hull now
argues for the donor class and the superstructure for the true class, and then we ask which the
classifier follows. It keeps the true (superstructure) class $73\%$ of the time and adopts the donor hull's
class only $4.8\%$ of the time, the remaining $22\%$ defaulting to a third class
(Fig.~\ref{fig:conflict}, top). Even in direct disagreement, the hull rarely overrides the
superstructure.

The same holds per class (Fig.~\ref{fig:conflict}, bottom). For each of the $23$ classes with a
defined hull-share (of the $25$ tested in the conflict study; two have an undefined hull-share) the
hull sensitivity exceeds the rate at which the hull is actually sufficient to flip the prediction,
and every such class sits well below the $y{=}x$ line. Substantial hull sensitivity therefore reflects
corroborating evidence rather than the hull overriding the class, and can be appropriate for vessel
types the hull genuinely defines. This is why the main text reports sensitivity as a measurement, kept separate from any
judgment of whether the reliance is warranted.

\section{Per-cell influence and per-seed audit}
Table~\ref{tab:cells} gives the full per-cell influence (raw cross-class, same-class
control, and corrected) underlying Fig.~\ref{fig:spectrum}a and the $54/46$
superstructure/hull split; the same-class column is the compositing floor that
Eq.~\eqref{eq:corr} subtracts. Table~\ref{tab:perseed} gives per-seed hull-share for the
audit methods, showing the across-seed consistency behind Table~\ref{tab:debias}.

\begin{table*}[t]
\centering\small
\begin{minipage}{0.52\linewidth}\centering
\caption{Per-cell influence ($277$ recipients; corrected hull-share $0.46$, $95\%$ CI $[0.41,0.52]$).}
\label{tab:cells}
\begin{tabular}{lccc}
\toprule
cell & cross $\Inf^{\times}$ & same $\Inf^{=}$ & corr.\ $\Inf^{\text{corr}}$ \\
\midrule
hull-fwd & $0.137$ & $0.092$ & $0.045$ \\
hull-mid & $0.211$ & $0.139$ & $0.072$ \\
hull-aft & $0.138$ & $0.090$ & $0.048$ \\
sup-fwd  & $0.150$ & $0.107$ & $0.043$ \\
sup-mid  & $0.284$ & $0.177$ & $0.107$ \\
sup-aft  & $0.153$ & $0.113$ & $0.040$ \\
\midrule
hull $\Sigma$ & & & $0.165\ (46\%)$ \\
sup $\Sigma$  & & & $0.190\ (54\%)$ \\
\bottomrule
\end{tabular}
\end{minipage}\hfill
\begin{minipage}{0.44\linewidth}\centering
\caption{Per-seed hull-share.}
\label{tab:perseed}
\begin{tabular}{lccc}
\toprule
method & s42 & s43 & s44 \\
\midrule
ERM       & $.393$ & $.398$ & $.392$ \\
CutMix    & $.401$ & $.383$ & $.405$ \\
hull-rand.& $.420$ & $.406$ & $.398$ \\
JTT       & $.422$ & $.400$ & $.375$ \\
LfF       & $.433$ & $.397$ & $.422$ \\
AFR       & $.397$ & $.410$ & $.389$ \\
bg-swap   & $.465$ & $.468$ & $.450$ \\
\bottomrule
\end{tabular}
\end{minipage}
\end{table*}

\medskip
\noindent Table~\ref{tab:phase2detail} gives the per-method controlled-shortcut numbers
(worst-group accuracy and interventional reliance) behind Fig.~\ref{fig:validation} and
Table~\ref{tab:replication}, for both the appearance-recolor principle (hull-color shortcut)
and the anatomical part-swap operator (pooled over three donor-hull shortcuts).

\begin{table*}[t]
\centering\small
\caption{\textbf{Controlled-shortcut per-method detail} (mean$\pm$std). Left: the
hull-color appearance shortcut ($3$ seeds). Right: the anatomical hull-graft operator,
pooled over $3$ donor-hull shortcuts $\times$ $3$ seeds ($9$ runs/method). Both track
worst-group accuracy (pooled $r{=}{-}0.97$ and $r{=}{-}0.941$).}
\label{tab:phase2detail}
\begin{minipage}{0.49\linewidth}\centering
{\footnotesize appearance recolor}\\[2pt]
\begin{tabular}{lcc}
\toprule
method & WGA $\uparrow$ & flip $\downarrow$ \\
\midrule
ERM      & $0.373{\scriptstyle\pm.038}$ & $0.389{\scriptstyle\pm.025}$ \\
SUBG     & $0.420{\scriptstyle\pm.017}$ & $0.335{\scriptstyle\pm.009}$ \\
JTT      & $0.410{\scriptstyle\pm.035}$ & $0.343{\scriptstyle\pm.071}$ \\
GroupDRO & $0.600{\scriptstyle\pm.010}$ & $0.196{\scriptstyle\pm.021}$ \\
DFR      & $\mathbf{0.730}{\scriptstyle\pm.038}$ & $\mathbf{0.120}{\scriptstyle\pm.044}$ \\
\bottomrule
\end{tabular}
\end{minipage}\hfill
\begin{minipage}{0.49\linewidth}\centering
{\footnotesize anatomical part-swap}\\[2pt]
\begin{tabular}{lcc}
\toprule
method & WGA $\uparrow$ & flip $\downarrow$ \\
\midrule
ERM      & $0.197{\scriptstyle\pm.054}$ & $0.563{\scriptstyle\pm.050}$ \\
SUBG     & $0.244{\scriptstyle\pm.119}$ & $0.504{\scriptstyle\pm.092}$ \\
JTT      & $0.250{\scriptstyle\pm.093}$ & $0.529{\scriptstyle\pm.062}$ \\
GroupDRO & $0.471{\scriptstyle\pm.080}$ & $0.269{\scriptstyle\pm.076}$ \\
DFR      & $\mathbf{0.653}{\scriptstyle\pm.050}$ & $\mathbf{0.123}{\scriptstyle\pm.041}$ \\
\bottomrule
\end{tabular}
\end{minipage}
\end{table*}

\section{De-biasing audit: the two axes}
\label{app:debias}
\begin{figure}[h]
\centering
\includegraphics[width=0.9\linewidth]{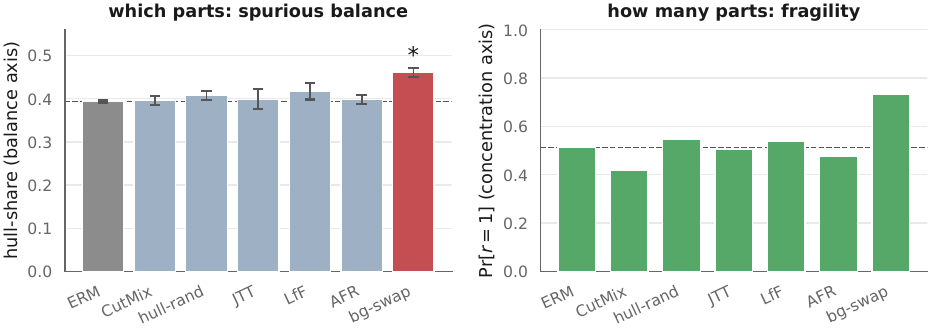}
\caption{\textbf{Two axes move independently} (visualizes Table~\ref{tab:debias},
Sec.~\ref{sec:debias}). Left: hull-share (which parts); $^{*}$ significant vs.\ ERM. Right:
$\Pr[r{=}1]$ (how many parts). CutMix lowers fragility (right) without lowering hull reliance
(left); bg-swap worsens both.}
\label{fig:debias}
\end{figure}

Part reliance moves along two independent axes (Sec.~\ref{sec:debias}, Table~\ref{tab:debias}), namely
which parts carry the decision (the hull-share) and how many parts must change to flip
it ($\Pr[r{=}1]$). Fig.~\ref{fig:debias} visualizes both. On the first axis (left) no group-label-free method
reduces the hull-share; the only significant mover, background-swap, raises it. The number of
parts moves separately (right): CutMix sharply lowers single-part fragility, spreading the decision
across cells, without making it any less biased toward the hull, so the two axes are decoupled, and an
augmentation can make a model more distributed while leaving its hull reliance untouched.

\medskip\noindent\textbf{Recipient selection.} Reliance is scored on each model's correctly classified
recipients, and each paired $\Delta$ vs.\ ERM is averaged over recipients that
both that method and ERM classify correctly. This is necessary (a flip rate is undefined on an image the model
already misclassifies), but it conditions the measured population on model success, so a de-biaser
can shift both its reliance and which recipients qualify. The paired both-correct design mitigates
this by comparing on a shared set. We also run the cleaner variant directly: a fixed-recipient audit
that scores every method on the identical $104$-recipient pool (no correct-only filter) using a
continuous response, the true-class probability $P(y\mid\phi(x,d,z))$, which is defined even when the
original prediction is wrong; the soft cell influence is the mean $|P(y\mid z)-P(y\mid z\oplus e_p)|$.
On this fixed population the soft hull-share is again flat across the group-label-free methods (paired
change vs.\ ERM from $-0.000$ to $+0.021$, none a reduction; CutMix $+0.006$, JTT $+0.001$, LfF
$+0.021$, AFR $-0.000$, hull-randomization $+0.003$), and only background-swap moves ($+0.030$). The
null is therefore not an artifact of correct-only recipient selection.

\medskip\noindent\textbf{What the audit could detect (minimum detectable effect).} Because the
headline on the first axis is a null, it is only as strong as the audit's power. The comparison is a
paired change over three seeds, with seed-to-seed std of the paired delta between $0.006$ and
$0.019$ across methods. For a paired $t$-test at $\alpha{=}0.05$ (two-sided, two degrees of freedom) and
$80\%$ power, this gives a minimum detectable effect of about $0.02$--$0.06$ per method and $\approx0.04$
pooled. Background-swap's $+0.069$ clears this comfortably and is flagged significant, whereas every
group-label-free delta ($\le0.022$) sits below it. The null therefore rules out any group-label-free
reduction above ${\sim}0.04$, not reductions of every size; separating a genuine tiny reduction from
none would need more seeds or recipients.

\medskip\noindent\textbf{Relative vs.\ absolute influence.} Hull-share is a ratio, so an unchanged
value could in principle hide a proportional drop in \emph{absolute} influence. It does not here.
Table~\ref{tab:absinf} reports the mean absolute hull, superstructure, and total flip-influence per
method on the same cross-class cube. Across the group-label-free methods the paired change in
absolute hull influence vs.\ ERM spans $-0.008$ to $+0.031$ and the change in total influence
$-0.059$ to $-0.002$, all within seed noise (paired std $0.01$--$0.06$); only background-swap moves
significantly, and it \emph{raises} both. The unchanged hull-share therefore reflects genuinely
unchanged absolute part sensitivity, not a proportional reduction masked by the ratio.

\begin{table}[h]\centering
\caption{\textbf{Absolute influence is flat, not just the ratio} (App.~\ref{app:debias}; $3$ seeds,
mean$\pm$std over seeds; same audit as Table~\ref{tab:debias}). Mean absolute hull, superstructure,
and total cross-class flip-influence per method, with the paired change vs.\ ERM in absolute hull and
total influence. No group-label-free method reduces absolute hull influence, and the small total
changes are within seed noise; only background-swap (positive control) moves, raising both.}
\label{tab:absinf}
\resizebox{\columnwidth}{!}{%
\begin{tabular}{lccccc}
\toprule
method & hull & sup.\ & total & $\Delta$hull & $\Delta$total \\
\midrule
ERM        & $0.461$ & $0.602$ & $1.063$ & --- & --- \\
CutMix     & $0.455$ & $0.552$ & $1.006$ & $-0.008{\scriptstyle\pm.031}$ & $-0.059{\scriptstyle\pm.058}$ \\
hull-rand. & $0.468$ & $0.562$ & $1.030$ & $+0.006{\scriptstyle\pm.027}$ & $-0.032{\scriptstyle\pm.044}$ \\
JTT        & $0.463$ & $0.580$ & $1.042$ & $-0.002{\scriptstyle\pm.034}$ & $-0.022{\scriptstyle\pm.034}$ \\
LfF        & $0.494$ & $0.567$ & $1.061$ & $+0.031{\scriptstyle\pm.041}$ & $-0.006{\scriptstyle\pm.054}$ \\
AFR        & $0.460$ & $0.590$ & $1.050$ & $+0.005{\scriptstyle\pm.013}$ & $-0.002{\scriptstyle\pm.012}$ \\
bg-swap    & $\mathbf{0.592}$ & $0.621$ & $\mathbf{1.214}$ & $\mathbf{+0.143}{\scriptstyle\pm.046}$ & $\mathbf{+0.186}{\scriptstyle\pm.042}$ \\
\bottomrule
\end{tabular}}
\end{table}

\section{Comparison to correlational attribution}
\label{app:attribution}
Does standard attribution already expose this reliance? On a controlled hull-color run (the
shortcut of Sec.~\ref{sec:phase2}, where the hull is tinted red/blue and correlated with the label,
so the ground-truth shortcut region is the hull mask), we score three attribution families,
Grad-CAM~\cite{gradcam} (activation), Integrated Gradients~\cite{ig} (gradient), and
RISE~\cite{rise} (perturbation), by the fraction of attribution mass they place on the
planted hull region, for five de-biasers over three seeds ($120$ correct test images each;
hull area~${\approx}0.12$). All three locate the shortcut, with Grad-CAM and Integrated Gradients
placing mass well above the hull area and RISE at about the hull area. Their magnitude is
nearly flat across de-biasers (range~${\le}0.05$;
Table~\ref{tab:attribution}). Our interventional flip on the same models spans $0.27$,
roughly $5{\times}$ more. As a model de-biases, correlational attribution barely moves while
the flip collapses. (Small residual changes in Grad-CAM mass can even correlate with
worst-group accuracy, but the effect is an order of magnitude too small to indicate
how much reliance changed.) Attribution answers where a model looks; only the
intervention answers how much the decision depends on the part.

\begin{table}[t]\centering
\caption{\textbf{Correlational attribution locates the shortcut but cannot measure reliance.}
Fraction of attribution mass on the planted hull region (controlled hull-color shortcut, $5$
de-biasers $\times$ $3$ seeds, $120$ images each; hull area~${\approx}0.12$). Every method
stays nearly flat across de-biasers (range~${\le}0.05$), whereas our interventional
hull-color flip spans $0.27$ over the same models, so attribution reports
where, not how much.}
\label{tab:attribution}
\begin{tabular}{lccc}
\toprule
de-biaser & Grad-CAM & IG & RISE \\
\midrule
ERM      & $0.34$ & $0.21$ & $0.12$ \\
SUBG     & $0.32$ & $0.18$ & $0.12$ \\
JTT      & $0.36$ & $0.23$ & $0.12$ \\
GroupDRO & $0.33$ & $0.18$ & $0.12$ \\
DFR      & $0.31$ & $0.20$ & $0.12$ \\
\midrule
\textbf{range across de-biasers} & $0.05$ & $0.05$ & $0.00$ \\
\bottomrule
\end{tabular}
\end{table}

\section{Beyond maritime: Stanford Cars}
\label{app:cars}
This appendix expands the non-maritime instantiation of Sec.~\ref{sec:cars}. It covers how the part
lattice is built for a domain with a different anatomy, the natural task part reliance readout, and the
convergent-validity study in full.

\begin{figure}[t]
\centering
\includegraphics[width=\linewidth]{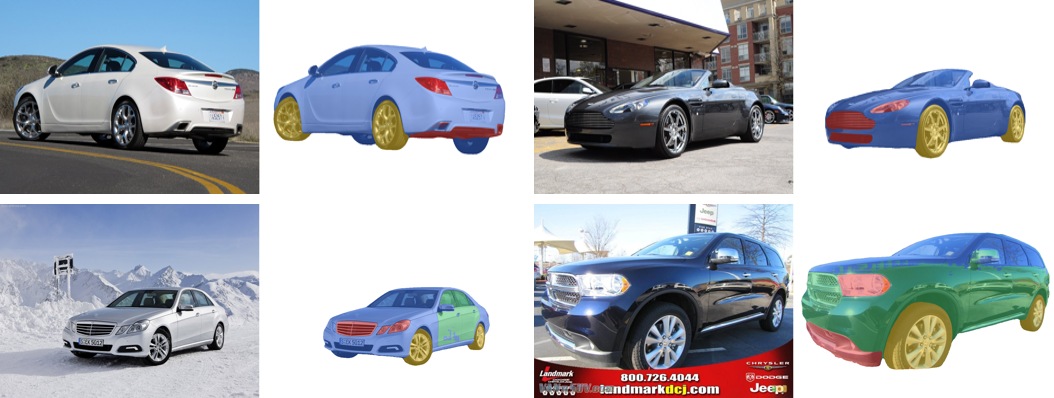}
\caption{\textbf{A non-maritime instantiation} (Sec.~\ref{sec:cars}). Four Stanford Cars
of distinct body styles (\emph{left} of each pair) and their natural $5$-part decomposition
(\emph{right}): \{body (gray), front-face (red), greenhouse (blue), body-side (green), wheels
(yellow)\}. Parts come from an open-vocabulary GroundedSAM prompt and the geometric PCA pose; the
hull/superstructure segmenter and the forward/mid/aft grid are not used. Over the resulting
$2^5$-swap cube the greenhouse (roof and glazing) carries the majority of corrected
interventional influence, well above the front grille.}
\label{fig:cars}
\end{figure}

\medskip\noindent\textbf{Part extraction and registration.} A car has no hull or superstructure, so
neither the maritime segmenter nor the forward/mid/aft geometry applies. Instead we prompt an
open-vocabulary GroundedSAM~\cite{groundingdino,sam} detector (box threshold $0.15$, text
threshold $0.12$) with the seven-token part prompt
``\texttt{wheel. windshield. headlight. grille. car door. bumper. car roof.}'' and
resolve the detections into a five-way partition \{body, front-face, greenhouse, body-side, wheel\}
by a fixed token-to-part map: \texttt{wheel}$\rightarrow$wheel; \texttt{windshield},
\texttt{car roof}$\rightarrow$greenhouse; \texttt{headlight}, \texttt{grille},
\texttt{bumper}$\rightarrow$front-face; \texttt{car door}$\rightarrow$body-side; and \emph{body}
is the residual backbone (every object pixel not claimed by a named part). Detections are painted
in ascending confidence so the higher-scoring detection wins any overlap, and a part is kept only
if its mask exceeds $0.3\%$ of the object mask. A car is accepted iff at least three parts are
named and they cover $\ge45\%$ of the object (with object area fraction $\in[0.04,0.93]$ and PCA
aspect $\in[1.35,6.0]$ to reject head-on/rear views). The object is then registered exactly as a
vessel is, by a closed-form similarity transform read
from its PCA pose (no trained aligner). Nothing is retrained, and the only learned component is the
frozen classifier under audit. Applied to the Stanford Cars~\cite{stanfordcars} test split (we
sample up to a fixed number of cars per model) this
yields a clean decomposition for $759$ of $784$ sampled cars ($97\%$), covering all $196$ models with
$\ge2$ instances each (Fig.~\ref{fig:cars}).

\begin{figure}[t]
\centering
\includegraphics[width=\linewidth]{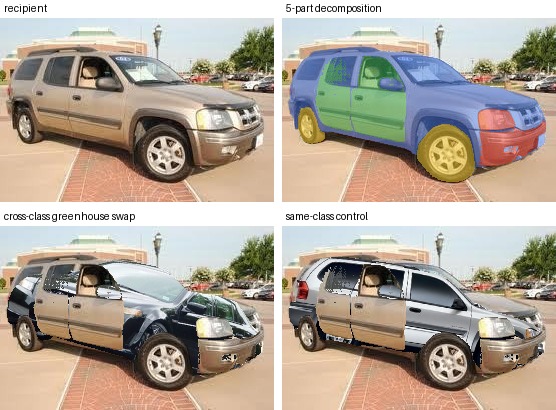}
\caption{\textbf{The car swap operator} (Sec.~\ref{sec:cars}). A recipient car, its natural
$5$-part decomposition, and a cross-class greenhouse swap that flips the predicted class, shown
against the label-preserving same-class control that isolates the compositing artifact. This is the
car analogue of the vessel operator in Fig.~\ref{fig:qual}.}
\label{fig:cars-swap}
\end{figure}

\begin{figure}[t]
\centering
\includegraphics[width=\linewidth]{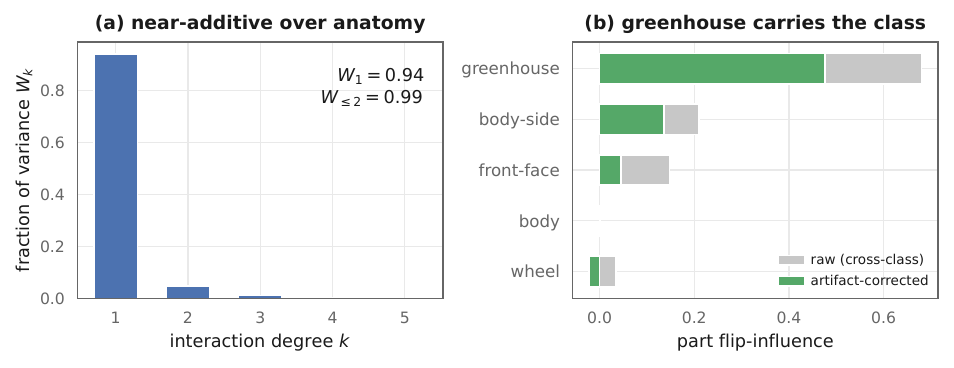}
\caption{\textbf{Cars: nearly additive anatomy and greenhouse dominance} (App.~\ref{app:cars}).
\emph{(a)} Walsh variance by interaction degree over the $2^5$ cube: main effects hold $W_1{=}0.94$
and degree-$\le2$ holds $0.99$, so per-part attribution captures almost all of the decision.
\emph{(b)} Per-part flip-influence, raw (gray) and artifact-corrected (green): the greenhouse
dominates, while the wheels and body fall to $\approx0$ once the same-class control adjusts for the
compositing floor.}
\label{fig:cars-spectrum}
\end{figure}

\medskip\noindent\textbf{Natural task part reliance.} We train a ConvNeXt-Tiny to $92.4\%$ test
accuracy and audit $80$ correctly-classified recipients over the resulting $2^5{=}32$-swap cube.
Recognition is nearly additive, even more sharply than on vessels, with single-part main effects
holding $W_1{=}0.94$ of the configuration variance and degree-$\le2$ holding $0.99$
(Fig.~\ref{fig:cars-spectrum}a), and $95\%$ of cars flip on one swap ($r{=}1$: one realistic
part swap flips the class). The artifact-corrected influence
(Table~\ref{tab:cars-influence}, Fig.~\ref{fig:cars-spectrum}b) concentrates on the
greenhouse (the roof and glazing) at corrected influence $0.48$, the top
part in $58/80$ cars and $3.5\times$ the next (body-side, $0.14$). The front-face contributes
little ($0.04$), and the wheels and body reduce to $\approx0$ under the same-class control, as
class-agnostic parts should.

\begin{table}[t]\centering
\caption{\textbf{Cars: per-part flip-influence} ($80$ recipients, $2^5$ cube; App.~\ref{app:cars}).
Raw cross-class $\Inf^{\times}$, same-class control $\Inf^{=}$, and corrected $\Inf^{\text{corr}}$.
The greenhouse carries the majority of corrected influence; wheels and body reduce to $\approx0$
under the control, as class-agnostic parts should.}
\label{tab:cars-influence}
\begin{tabular}{lccc}
\toprule
part & cross $\Inf^{\times}$ & same $\Inf^{=}$ & corr.\ $\Inf^{\text{corr}}$ \\
\midrule
greenhouse & $0.680$ & $0.205$ & $\mathbf{0.476}$ \\
body-side  & $0.210$ & $0.073$ & $0.137$ \\
front-face & $0.148$ & $0.103$ & $0.045$ \\
body       & $0.000$ & $0.000$ & $0.000$ \\
wheel      & $0.034$ & $0.056$ & $-0.022$ \\
\bottomrule
\end{tabular}
\end{table}

\begin{figure}[t]
\centering
\includegraphics[width=0.78\linewidth]{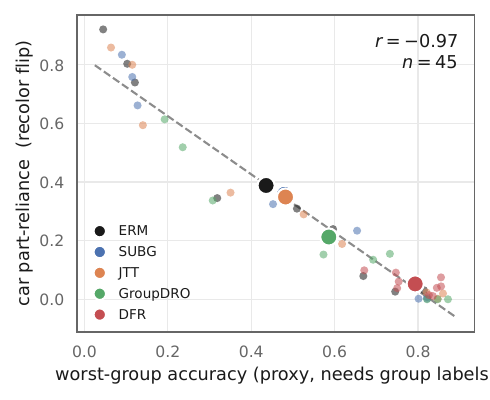}
\caption{\textbf{Convergent validity ports to cars} (Sec.~\ref{sec:cars}). Three graded color
shortcuts ($3\times3$ seeds $\times5$ de-biasers, $45$ configs): group-label-free recolor-flip
reliance tracks worst-group accuracy across the full range (pooled $r{=}{-}0.97$; $-0.97$ without
DFR), ERM leaning on the shortcut and DFR removing it. The car swap operator is shown in
Fig.~\ref{fig:cars-swap}.}
\label{fig:carsconv}
\end{figure}

\medskip\noindent\textbf{Convergent validity.} We repeat the controlled-shortcut validation of
Sec.~\ref{sec:phase2} on the car lattice. Three graded color shortcuts are planted on the car's
parts and correlated with a binary model-group label (balanced at test). The three are a strong \emph{painted}
shortcut (all painted parts recolored), a medium \emph{background} shortcut, and a weak corner
\emph{watermark} specificity control. Across five de-biasers $\times$ three seeds $\times$ three
shortcuts ($45$ configurations) the group-label-free recolor-flip reliance tracks worst-group
accuracy at pooled $r{=}{-}0.973$, and the graded strengths spread the configurations across the
full accuracy range (Fig.~\ref{fig:carsconv}, Table~\ref{tab:cars-convergent}). ERM leans on each shortcut and DFR removes it,
exactly as on vessels. The correlation is not an artifact of DFR's extreme position, since excluding
DFR entirely leaves it unchanged ($r{=}{-}0.973$; Spearman $\rho{=}{-}0.94$). The watermark is too small
to be learned (ERM reliance $\le0.05$), and the measure correctly reports near-zero reliance for it.

\begin{table}[t]\centering
\caption{\textbf{Cars convergent validity, per shortcut} (ResNet-50, $3$ seeds each;
App.~\ref{app:cars}). Three graded shortcuts span the worst-group-accuracy range; for each, WGA rises
and the group-label-free recolor-flip reliance falls from ERM to DFR. Pooled $r{=}{-}0.973$ over all
$45$ configurations ($-0.973$ excluding DFR).}
\label{tab:cars-convergent}
\resizebox{\columnwidth}{!}{%
\begin{tabular}{llccc}
\toprule
shortcut & strength & ERM (WGA/flip) & DFR (WGA/flip) & corr \\
\midrule
painted    & strong & $0.09/0.82$ & $0.72/0.08$ & $-0.962$ \\
background & medium & $0.47/0.30$ & $0.82/0.05$ & $-0.923$ \\
watermark$^{\dagger}$ & weak & $0.74/0.05$ & $0.84/0.02$ & $-0.743$ \\
\bottomrule
\end{tabular}}
\end{table}

\medskip\noindent\textbf{What porting HullProbe requires.} Table~\ref{tab:porting} lists, for the two
instantiations in this paper, every component a new domain must supply. Only two are domain-specific:
an exhaustive partition of the object into a handful of named parts that every instance has, and a
closed-form canonical frame in which part $p$ of one instance corresponds to part $p$ of another.
Everything downstream (swap operator, same-class control, cube evaluation, Walsh readouts) is reused
unchanged, and the classifier is queried as a black box. The decisive constraint is registration.
Elongated, rigid objects photographed from a consistent viewpoint admit a PCA frame (vessels and
side-view cars here; plausibly aircraft or trains, which we have not tested); articulated,
deformable, or viewpoint-varying objects (animals, people, most ImageNet categories) do not, and
would need learned keypoints or 3D correspondence before the swap is anatomically meaningful. The cube also grows as $2^{|P|}$, so exhaustive evaluation is practical
up to roughly eight parts.

\begin{table*}[t]\centering\small
\caption{\textbf{What it takes to port HullProbe} (App.~\ref{app:cars}). Components a new domain must
supply, as instantiated for vessels and cars. Only the part decomposition and the registration are
domain-specific; the swap operator, artifact control, cube evaluation, and readouts are shared.}
\label{tab:porting}
\renewcommand{\arraystretch}{1.15}
\begin{tabular}{@{}p{0.17\linewidth}p{0.25\linewidth}p{0.25\linewidth}p{0.27\linewidth}@{}}
\toprule
component & vessels (MARVEL, SeaShips) & cars (Stanford Cars) & requirement for a new domain \\
\midrule
part set $P$ & $2{\times}3$: \{hull, sup\} $\times$ \{fwd, mid, aft\}, $|P|{=}6$ & 5 named parts, $|P|{=}5$ & exhaustive partition into $\lesssim8$ parts present on every instance \\
part extractor & SegFormer-B2 fine-tuned on $1{,}014$ masks (hull/sup); PCA thirds & GroundedSAM prompt + fixed token-to-part map; no annotation & a learned or promptable segmenter, or a geometric split \\
registration & closed-form PCA similarity transform & same & a canonical frame with part-to-part correspondence (rigid, elongated, consistent view) \\
validity filter & MARVEL: PCA aspect $\ge1.40$; SeaShips: every cell $\ge150$px ($76\%$ yield) & $\ge3$ parts covering $\ge45\%$, aspect $\in[1.35,6]$ ($97\%$ yield) & a rule rejecting views where parts are missing or unregistrable \\
donor pool & $K{=}4$ aspect-compatible cross-class $+\,2$ same-class per recipient & same & $\ge3$ held-out instances per class \\
classifier access & black box, predicted labels (probabilities for the soft variant) & same & no gradients, retraining, or group labels \\
compute per recipient & $64{\times}6{=}384$ forward passes, a few seconds on one GPU (compositing-bound) & $32{\times}6{=}192$ forward passes & $2^{|P|}(K_{\times}{+}K_{=})$ forward passes \\
\bottomrule
\end{tabular}
\end{table*}

\section{Zero-shot transfer to SeaShips}
\label{app:seaships}
\begin{figure*}[t]
\centering
\includegraphics[width=\linewidth]{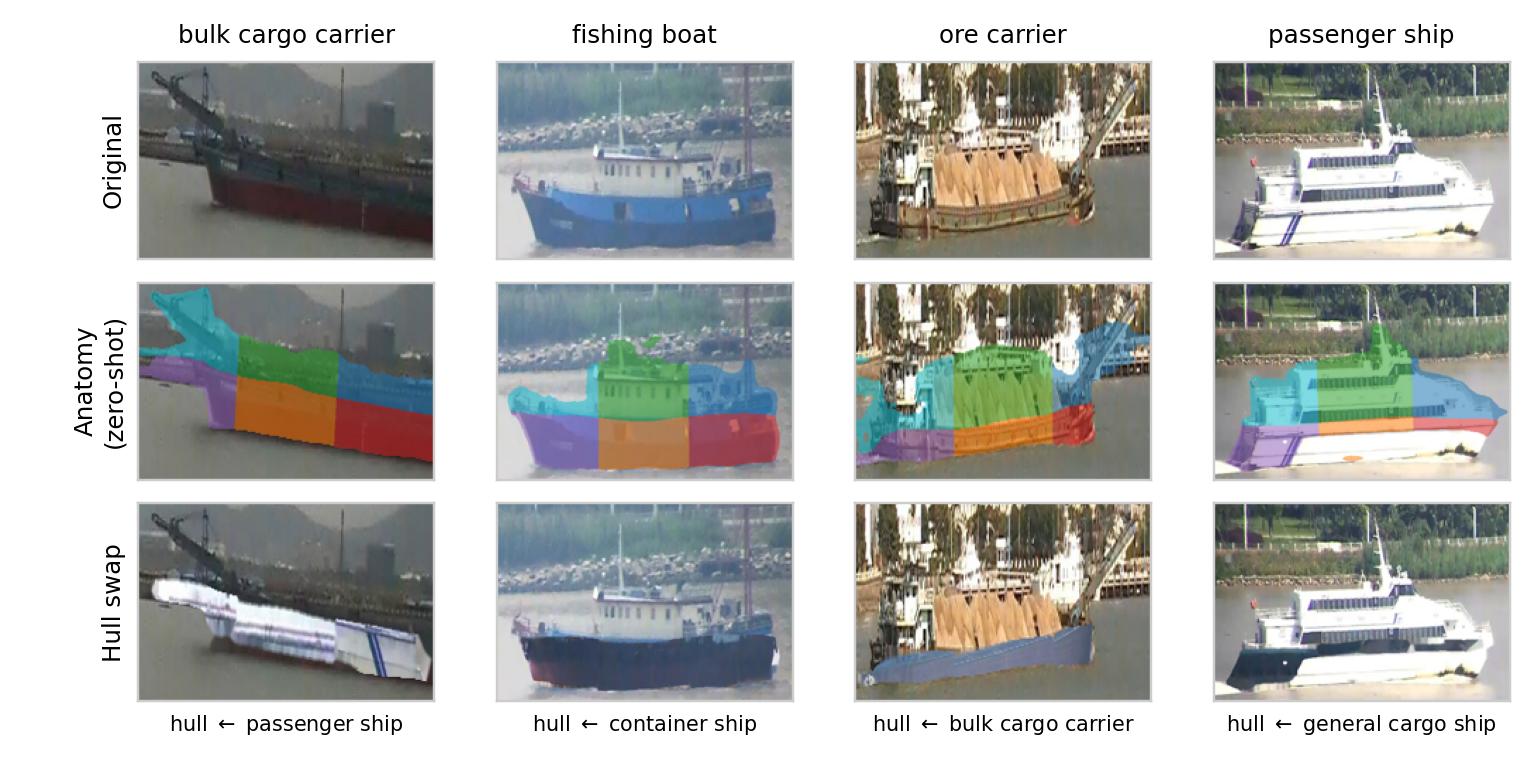}
\caption{\textbf{Zero-shot transfer to SeaShips (second dataset).} Both the MARVEL-trained
SegFormer segmenter (middle, the hull/superstructure$\times$forward/mid/aft
decomposition) and the swap operator (\emph{bottom}, a cross-class hull swap, \emph{hull
$\leftarrow$ donor type}) apply to SeaShips vessels with no retraining, staying clean
across distinct ship types. This is the qualitative counterpart to the numerical replication
(Sec.~\ref{sec:generality}). Segmentation is zero-shot, so composites are noisier than on
MARVEL, which the same-class control absorbs.}
\label{fig:seaships}
\end{figure*}

The maritime findings replicate on a second, independent dataset. SeaShips~\cite{seaships} is a
detection dataset of $6$ ship types collected by different cameras; we crop its vessels, apply the
MARVEL-trained SegFormer segmenter zero-shot (no retraining), and train a fresh classifier to
$98.4\%$ accuracy. SeaShips yields $4{,}086$ crops with a valid decomposition ($76\%$), the pool from which recipients
and donors are drawn. Over a class-balanced $n{=}48$ recipients from that pool, the low-order
structure not only holds but strengthens, with degree-$\le2$ variance rising to $0.91$ (vs.\
$0.80$ on MARVEL), hull reliance staying substantial (hull-share $0.31$), and the
semantic-$L_0$ tightening ($25\%$ flip on one swap vs.\ $44\%$), tracking the higher clean accuracy.
The segmenter is zero-shot and $n{=}48$ recipients is small, so these numbers are indicative; still,
nearly additive anatomy holding on both datasets and all three architectures despite a
large accuracy gap points to a real structural regularity rather than a MARVEL-specific artifact.
Fig.~\ref{fig:seaships} shows that both the segmenter and the swap operator stay clean on SeaShips
vessels with no retraining.

\section{Fourier--Walsh background}
\label{sec:fw-appendix}
For $g:\{0,1\}^n\!\to\!\mathbb{R}$ under the uniform measure, the parities
$\chi_S(z)=\prod_{p\in S}(-1)^{z_p}$ form an orthonormal basis because
$\mathbb{E}_z[\chi_S\chi_T]=\prod_{p\in S\triangle T}\mathbb{E}[(-1)^{z_p}]$ and
$\mathbb{E}[(-1)^{z_p}]=0$; hence it is $1$ iff $S{=}T$. Thus $\hat g(S)=\langle
g,\chi_S\rangle$: the main effects are the degree-1 coefficients
($\hat g(\{p\})=\tfrac12 M_p$ from Eq.~\eqref{eq:main}, by conditioning on $z_p$) and the
pairwise interactions are the degree-2 coefficients ($\hat g(\{p,q\})=\tfrac14 I_{pq}$ from
Eq.~\eqref{eq:inter}, by conditioning on $z_p,z_q$), so the degree-$k$ mass $V_k$ below
aggregates exactly these order-$k$ terms. Parseval gives the exact variance decomposition
\begin{equation}
\operatorname{Var}_z[g]=\sum_{S\ne\emptyset}\hat g(S)^2,\qquad
V_k=\!\!\sum_{|S|=k}\!\!\hat g(S)^2,
\label{eq:parseval}
\end{equation}
so $W_k{=}V_k/\!\operatorname{Var}_z[g]$ is the fraction of behavior at interaction order $k$; it
is this normalized $W_k$ that the main text and figures report (e.g.\ $W_1{=}0.63$). For
each Boolean $F_d$, the per-donor influence in Eq.~\eqref{eq:influence} equals
$\sum_{S\ni p}\hat F_d(S)^2$ via the
discrete derivative, so total influence $\sum_p\Inf_p=\mathbb{E}_d\sum_S|S|\hat F_d(S)^2$ is the
average sensitivity of the decision to part swaps.

\section{Anatomy extraction and the swap operator}
\label{sec:operator-appendix}
How the six cells are obtained and how a cell is swapped. Only the hull/superstructure
segmenter is learned; the longitudinal split and all registration are closed-form from
the vessel's PCA pose, with no trained aligner.

\textbf{Vessel silhouette (zero-shot).} The vessel mask is the boat/ship prediction of an
ADE20K-pretrained SegFormer-B2~\cite{segformer,ade20k}, used without fine-tuning (summed boat and
ship class probability $\ge0.12$). It fixes the silhouette, the PCA pose, and the outer boundary of
every cell.

\textbf{Hull/superstructure (learned).} An annotator placed the hull/superstructure boundary of
$1{,}014$ vessels as a deck line inside the silhouette, starting from a geometric estimate and
shifting or rotating it as needed ($88\%$ of the labels differ from that estimate, by a median
$17\%$ of vessel pixels; $214$ further candidates were rejected). We fine-tune SegFormer-B2 from the
same ADE20K initialization on these labels to a three-class map \{background, hull,
superstructure\}, reaching mean hull/superstructure IoU $0.756$ on its $202$ held-out vessels, and
run it once over all $34{,}768$ vessels. Inside the silhouette the superstructure is what the
segmenter labels as such and the hull is the rest, so the two touch and are jointly complete.
App.~\ref{app:gtmask} measures how the segmenter's errors propagate into the audit.

\textbf{Longitudinal thirds (geometric).} We fit PCA to the vessel mask's pixel covariance; the
leading eigenvector is the longitudinal axis and the pixel mean is the origin. The eigenvector
sign follows a fixed image-space convention, oriented to point in the $+x$ (image-right)
direction (negated if its $x$-component is negative), so the low-projection end is a consistent
canonical ``forward'' third across vessels. We do not detect the physical bow; the fwd/mid/aft
labels are thirds of the canonical frame, and cross-vessel cell correspondence is by this shared frame
rather than by a semantic bow/stern cue (vessels in MARVEL are predominantly captured in a
consistent broadside orientation, which this convention relies on). Each vessel pixel gets a
signed longitudinal coordinate $L=(x{-}c_x)\cos\theta+(y{-}c_y)\sin\theta$, and
$[L_{\min},L_{\max}]$ is split into equal thirds to give fwd/mid/aft. This is deterministic (no
model is trained), and its product with \{hull, sup\} gives the six cells, which exactly partition
the vessel.

\textbf{Canonical frame.} Each vessel is mapped to a shared frame by a similarity transform $S$
read directly from its pose. It derotates by $\theta$ and translates the centroid, then scales so the
longitudinal extent $L_{\max}{-}L_{\min}$ becomes a fixed canonical width (deck line aligned to
a fixed row). $S$ is closed-form; again nothing is learned.

\textbf{Per-cell swap.} To replace recipient cell $p$ with a donor's cell $p$ we do not
paste a globally-warped donor (which lands the wrong donor region in the cell). Instead each
donor cell is fit to its matching recipient cell independently by a diagonal affine in the canonical frame, matching their longitudinal extent (the along-axis scale $a$), area (the determinant,
giving the cross-axis scale $d{=}\det/a$), and centroid (the translation), clamped to
$a{\in}[0.35,2.5]$ and $\det{\in}[0.20,4.0]$ to prevent degenerate warps. The donor's cell-$p$
texture is warped by this fit, stenciled onto the recipient's exact cell support (so the
composite silhouette is always the recipient's, with no overhang or holes), retained cells are
kept pixel-exact, and any residual gap is inpainted (OpenCV Telea, radius $3$). Thus
$z{=}(1,1,1,0,0,0)$ swaps all three hull cells.

\textbf{Why no learned aligner, and the artifact control.} Because the two decompositions
already register every vessel into one canonical PCA frame, cell-to-cell correspondence is
closed-form and a learned graft aligner proved unnecessary. Compositing still leaves
seams and inpainting; the same-class control (Eq.~\eqref{eq:corr}) repeats the identical
operation with a same-class donor and subtracts its effect, adjusting for these artifacts
under the class-independence assumption.

\section{Baseline and de-biasing methods}
\label{sec:methods-appendix}
All natural task methods (Sec.~\ref{sec:debias}, Table~\ref{tab:debias}) share one matched training
protocol (ConvNeXt-Tiny from ImageNet initialization, AdamW (lr $2{\times}10^{-4}$, weight
decay $0.05$), cosine schedule, $12$ epochs, batch $64$, class-balanced $250$ images/class,
cross-entropy with label smoothing $0.05$) and differ only in the objective or augmentation:
\begin{itemize}[leftmargin=1.2em,itemsep=1pt,topsep=1pt]
\item \textbf{ERM:} the untreated baseline (plain cross-entropy).
\item \textbf{CutMix}~\cite{cutmix}: with probability $0.5$ per batch a rectangular region of
another image is pasted in and the label mixed in proportion to the pasted area, a strong
region-mixing regularizer.
\item \textbf{Hull-randomization:} our anatomy augmentation, in which with probability $0.5$ the hull
is replaced by a shape-registered donor hull while the superstructure and label are kept; the
label-preserving counterpart of the intervention operator.
\item \textbf{Background-swap:} with probability $0.5$ the vessel is composited onto another
image's background (vessel and label kept), encouraging background invariance.
\item \textbf{JTT}~\cite{jtt} (Just Train Twice): train an identification ERM for $5$ epochs,
collect the training examples it misclassifies, and retrain from scratch with that error set
upweighted $\times20$, a group-free approximation to worst-group training.
\item \textbf{LfF}~\cite{lff} (Learning from Failure): jointly train a biased network $f_B$
with generalized cross-entropy ($q{=}0.7$, which amplifies easy/spurious cues) and a debiased
network $f_D$ whose per-example loss is weighted by the relative difficulty
$W{=}L_B/(L_B{+}L_D)$ (EMA-smoothed, $\alpha{=}0.7$); the audited model is $f_D$.
\item \textbf{AFR}~\cite{afr} (Automatic Feature Reweighting): freeze the ERM features and
retrain only the last linear layer with per-example weights $(1-p_y)^{\gamma}$ ($\gamma{=}2.0$,
$p_y$ the ERM's confidence in the true class), upweighting the examples the ERM classifies with low
confidence, the group-free counterpart of DFR.
\end{itemize}
The controlled-shortcut studies (Sec.~\ref{sec:phase2}, Table~\ref{tab:phase2detail};
ResNet-50, three seeds) additionally evaluate group-supervised methods, which require
the planted group labels:
\begin{itemize}[leftmargin=1.2em,itemsep=1pt,topsep=1pt]
\item \textbf{SUBG}~\cite{subg}: subsample to equalize the four (label\,$\times$\,attribute)
groups, then ERM, a simple but strong balancing baseline.
\item \textbf{GroupDRO}~\cite{groupdro}: minimize the worst-group loss through an
online exponentiated reweighting over groups (step size $\eta{=}0.01$).
\item \textbf{DFR}~\cite{dfr} (Deep Feature Reweighting): freeze the ERM features and retrain
the last layer on a group-balanced set; AFR above is its group-free analogue.
\end{itemize}

\medskip\noindent\textbf{Classifier and segmenter training protocols.} The controlled-shortcut methods
(Table~\ref{tab:phase2detail}) fine-tune an ImageNet-pretrained ResNet-50 with AdamW (lr
$1{\times}10^{-4}$, weight decay $10^{-4}$), batch $64$, $8$ epochs; DFR/AFR retrain only the final
linear layer (lr $10^{-3}$/$10^{-2}$, weight decay $10^{-3}$) on frozen features. The natural task
attribution model is the same ConvNeXt-Tiny ERM protocol above ($77.6\%$). The cars classifier
($92.4\%$) reuses the maritime baseline protocol (ConvNeXt-Tiny, AdamW lr $2{\times}10^{-4}$, weight
decay $0.05$, cosine schedule, $12$ epochs, batch $64$, label smoothing $0.05$). The
hull/superstructure SegFormer-B2 is fine-tuned from ADE20K for $45$ epochs (batch $8$, lr
$6{\times}10^{-5}$, horizontal flips) on the $1{,}014$ annotations under a seeded $80/20$ train/val
split ($812/202$), keeping the epoch-$33$ checkpoint with the best validation mean IoU, $0.756$
(hull and superstructure IoU averaged, computed at $512$px over the full image). For the cars convergent-validity study, the
three graded shortcuts are planted with recolor opacity $\alpha$ and train-time
attribute--label alignment $p_{\text{align}}$: painted (strong) $\alpha{=}0.80$,
$p_{\text{align}}{=}0.95$; background (medium) $\alpha{=}0.75$, $p_{\text{align}}{=}0.90$; corner
watermark (weak specificity control) $\alpha{=}1.00$, $p_{\text{align}}{=}0.85$; all balanced at
test.

\section{Benchmark protocol}
\label{app:protocol}
HullProbe is an evaluation protocol layered on existing public datasets (MARVEL, SeaShips, and
Stanford Cars); our contribution is the measure and the artifacts that make it reproducible, not new
imagery. We release the evaluation harness, the fine-tuned SegFormer segmenter and its $1{,}014$
annotations, the composition code, the fixed recipient and donor identity lists, and every per-model
output behind our results at \releaseurl, so that anyone with the public source images can regenerate
the exact swap cubes. Because the interventional measure
depends on the intervention distribution, a HullProbe score is
comparable across models only if the protocol is fixed. The release therefore freezes every choice
that affects the number, so a future model is scored by classification alone.
\begin{itemize}[leftmargin=1.2em,itemsep=1pt,topsep=1pt]
\item \textbf{Recipients and donors.} Fixed identity lists per analysis (the $277$-recipient
attribution pool, the de-biasing pool, and the controlled-shortcut pools), drawn once from the
held-out MARVEL test split with seed $42$ and published verbatim.
\item \textbf{Donor law $\pi$.} A fixed donor bank per recipient: $K{=}4$ cross-class donors,
the nearest in aspect ratio among $60$ cross-class candidates drawn under the fixed seed, and $2$
same-class controls drawn at random under the same seed. The published lists record every donor
identity, so the bank is reproduced exactly rather than re-drawn. The random-donor and
$K\in\{2,8\}$ variants (App.~\ref{app:robustness}) are provided for sensitivity analysis only and
are not part of the standard score.
\item \textbf{Segmentation and the recipient pool.} The released SegFormer outputs are the
canonical region maps. The pool is every MARVEL subset-B test-split vessel with a region map and a
valid PCA pose of aspect $\ge1.40$ (rejecting near-square top and oblique views), $4{,}395$ vessels.
Cells may be empty ($5.8\%$ of the pool has at least one, e.g.\ no forward superstructure); an
empty cell simply carries zero influence. The pool and the recipient lists are published, so no
re-segmentation is required and the score does not depend on a researcher's segmenter.
\item \textbf{Composites and seeds.} The composition code and fixed seeds regenerate the exact
$64$-cell cubes for every recipient$\times$donor pair from the public source images, so scoring a new
classifier reproduces our composites deterministically rather than requiring us to redistribute them.
\end{itemize}
Vessel recognition has surveillance
implications, but HullProbe is diagnostic (it audits already-trained models), not a
deployment system.

\end{document}